\documentclass[12pt]{amsart}
\usepackage{amsbsy,amssymb,amsmath,amsthm,amscd,amsfonts,latexsym,amstext,delarray,
amsmath,graphicx} 
\usepackage{qtree}
\usepackage[margin=.8in]{geometry}
\usepackage{color}
\usepackage[all]{xy}

\newtheorem{thm}{Theorem}[section]
\newtheorem{prop}[thm]{Proposition}
\newtheorem{cor}[thm]{Corollary}
\newtheorem{lem}[thm]{Lemma}

\newtheorem{defn}[thm]{Definition}
\newtheorem{rem}[thm]{Remark}
\newtheorem{ex}[thm]{Example}

\usepackage[normalem]{ulem}

\usepackage[]{mdframed}

\usepackage{bigstrut}

\numberwithin{equation}{section}

\def\Q{{\mathbb Q}}

\def\cB{{\mathcal B}}

\def\cI{{\mathcal I}}

\def\cK{{\mathcal K}}
\def\cL{{\mathcal L}}
\def\cM{{\mathcal M}}

\def\cO{{\mathcal O}}

\def\cR{{\mathcal R}}
\def\cS{{\mathcal S}}
\def\cT{{\mathcal T}}

\def\cV{{\mathcal V}}

\def\bB{{\mathbb B}}

\def\fM{{\mathfrak M}}
\def\fT{{\mathfrak T}}
\def\fF{{\mathfrak F}}
\def\fO{{\mathfrak O}}

\title[Theta Theory]{Theta Theory: operads and coloring}
\author[M.Marcolli, R.K.Larson]{Matilde Marcolli \& Richard Larson}
\date{February, 2025}

\address{Department of Mathematics and Department of Computing and Mathematical Sciences, 
California Institute of Technology, CA 91125, USA}
\email{matilde@caltech.edu}
\address{Department of Linguistics, Stony Brook University, NY 11794-4376, USA}
\email{richard.larson@stonybrook.edu}

\begin{document}
\maketitle

\begin{abstract}
We give an explicit construction of the generating set of a colored operad that
implements theta theory in the mathematical model of Minimalism, in the form
of a coloring algorithm for syntactic objects. We show that the coproduct
operation on workspaces allows for a recursive implementation of the theta
criterion. We also show that this filtering by coloring rules on structures freely formed
by Merge is equivalent to a process of structure formation by a colored version
of Merge: the form of the generators of the colored operad then implies the dichotomy
is semantics between External and Internal Merge, where Internal Merge only
moves to non-theta positions. 
\end{abstract}

\section{Introduction}

This text elaborates and refines the discussion in \cite{MCB} of theta theory in the context of the
mathematical formulation of Merge and the Strong Minimalist Thesis. The approach outlined in \cite{MCB}
proposes the use of the mathematical formalism of colored operads to incorporate theta theory in
the mathematical model of Merge. However, the structure of the colored operad required for this
purpose is not explicitly discussed in \cite{MCB}. The main goal of the present paper is to fully 
develop a formulation of theta theory in the setting of the mathematical Minimalism. 

The main idea here is that theta theory is a coloring algorithm that acts on the products of
the free symmetric Merge, by filtering out structures on the basis of incorrect coloring, 
instead of constraining the action of Merge itself. The coloring algorithm is determined by
two parts: an assignment of coloring rules, and a consistency checking. The first can be
formulated in terms of a generating system for a colored operad. It incorporates the
usual linguistic notions of theta roles, theta grid, external and internal argument, non-theta positions,
theta-hierarchies.
The second step, the consistency checking, uses the coproduct decomposition to eliminate the
non-theta positions via admissible cuts, and then implements the theta criterion, checking  
the matching of theta roles on the remaining quotient part (deletion quotient). 
There is a similar use of the coproduct decomposition in a tricoloring
algorithm that assigns the structure of head and complement, and that allows
for an optimality formalization of the FOFC: this will be discussed elsewhere. 
However, it is worth mentioning here that the filters acting in between the free structure
formation of Merge and the interfaces appear to be describable in terms of a
similar formalism of coloring algorithms. In the case we discuss here the coloring
and the resulting consistency checking are specifically designed to track the
assignment of theta roles and the validity of the theta criterion. 

An important aspect of theta theory is the dichotomy (sometimes called ``duality")
between Internal and External Merge, with the assignment of theta roles performed by
External Merge, while Internal Merge performs movement to non-theta positions. One of
the results of this paper is to show that this dichotomy {\em follows} from the construction
of the colored operad that accounts for theta role assignments and theta grids. In other
words, it is not an additional, independent property of theta theory and can be
{\em derived} from the coloring algorithm. This result is obtained by comparing
free structure formation by Merge followed by filtering by the coloring algorithm with 
an equivalent reformulation in terms of a colored version of the grafting operator 
$\cB$ that is used in the description of the Merge action. 

\section{Theta Theory and Mathematical Minimalism}

The most recent form of Minimalism, based on free symmetric Merge and the Strong Minimalist Thesis, as
articulated in Chomsky's \cite{ChomskyGK}, \cite{ChomskyGE}, \cite{ChomskyElements}, admits a 
mathematical formulation, developed by Marcolli, Chomsky, and Berwick in \cite{MCB}.
In Section~3.8.1 of \cite{MCB}, an approach to incorporating Theta Theory into this mathematical
model is presented and briefly discussed. The main idea articulated there is the use of colored
operads, and algebras over colored operads, to account for the matching of theta roles. 

A key part of the algebraic structure, described in Section~3.8.1 of \cite{MCB}, 
is the structure of algebra over an operad: 
the set of syntactic objects $\fT_{\cS\cO_0}$ (identified with
the set of non-planar binary rooted trees with leaves 
labelled by elements of the set $\cS\cO_0$ of lexical items and syntactic features) is
an algebra over the operad of (non-labelled) non-planar binary rooted trees. 
We recall the relevant mathematical notions in \S \ref{operadSec}.

\subsection{Operads and algebras over operads}\label{operadSec}

We first recall the notion of operads, which was introduced in \cite{May} in the context of
algebraic topology. The notion of operads describes composable collections of operations
with varying numbers of inputs and one output where the compositions are
the natural ones that plug outputs of one operation into inputs of the next.
A related notion of algebra over an operad describes sets of elements that
can serve as inputs for the operations in the operad, generating outputs in the
same set, namely sets with an action of the operations in the operad. 

\subsubsection{Operads}
An operad (in the category of sets) is a collection
$\fO=\{ \fO(n) \}$ of sets $\fO(n)$ consisting of operations $T \in \fO(n)$ with $n$
inputs and one output. The algebraic structure of $\fO$ is given by composition operations
\begin{equation}\label{operadcomp}
 \gamma: \fO(n) \times \fO(k_1)\times \cdots \times \fO(k_n) \to \fO(k_1+\cdots + k_n) 
\end{equation} 
that plug the single output of an operation in $\fO(k_j)$ into the $j$-th input of an
operation in $\fO(n)$, resulting in a new operation where all the inputs of the operation
in $\fO(n)$ are saturated and there are exactly $k_1+\cdots + k_n$ new inputs coming 
from the inputs of the $\fO(k_j)$. The composition operations \eqref{operadcomp} are
associative, namely
\begin{equation}\label{gammassoc}
\begin{array}{c} \gamma(\gamma(T,T_1,\ldots,T_n);T_{1,1},\ldots,T_{1,m_1},\ldots,T_{n,1},\ldots, T_{n,m_n})= \\  \gamma(T;\gamma(T_1;T_{1,1},\ldots,T_{1,m_1}),\ldots,\gamma(T_n;T_{n,1},\ldots, T_{n,m_n})) \, . \end{array} 
\end{equation}
The operad is unital if there is a unit operation ${\bf 1}\in \fO(1)$ satisfying the two identities
$\gamma({\bf 1};T)=T$ and $\gamma(T;{\bf 1},\ldots,{\bf 1})=T$. 

If an operad is unital, then the composition operations \eqref{operadcomp} are equivalent to
the insertion operations
\begin{equation}\label{insertions}
 \circ_i: \fO(n)\otimes \fO(m) \to \fO(n+m-1). 
\end{equation}
For $1\leq j \leq a$ and $b,c\geq 0$, with 
$X\in \fO(a)$, $Y\in \fO(b)$, and $Z\in \fO(c)$, these insertions are subject to the conditions
$$
 (X\circ_j Y)\circ_i Z = \left\{ \begin{matrix} (X\circ_i Z)\circ_{j+c-1} Y & 1\leq i < j \\
X\circ_j (Y\circ_{i-j+1} Z)& j\leq i < b+j \\
(X\circ_{i-b+1} Z)\circ_j Y & j+b \leq i \leq a+b-1.  \end{matrix} \right. 
$$
The composition laws $\gamma$ as in \eqref{operadcomp}, satisfying the associativity condition \eqref{gammassoc}, 
can be obtained from the insertions $\circ_i$ of \eqref{insertions} through 
$$
 \gamma(X, Y_1,\ldots, Y_n)=(\cdots (X\circ_n Y_n)\circ_{n-1} Y_{n-1}) \cdots \circ_1 Y_1). 
$$
These two descriptions of operads are equivalent in the unital case, but inequivalent in
the non-unital case, see \cite{Markl}: non--unital operads defined through the insertion
operations $\circ_i$ are also non--unital operads with the
composition operations $\gamma$, but the converse no longer holds.

An operad $\fO$ is symmetric if the composition operations are equivariant with
respect to the actions of symmetric groups by permutations. More precisely,
for permutations $\sigma_i\in \Sigma_{n_i}$ and $\sigma\in \Sigma_m$
in the symmetric groups on $n_i$ (respectively, $m$) elements, two identities hold. 
The first one is given by
\begin{equation}\label{symoperad1}
 \gamma(\sigma(T); T_{\sigma^{-1}(1)},\ldots, T_{\sigma^{-1}(m)}) =
\tilde\sigma(\gamma(T; T_1,\ldots, T_m)), 
\end{equation}
where $\tilde\sigma \in \Sigma_{n_1+\cdots + n_m}$ is the
permutation that splits the set of indices into blocks of $n_i$ indices and permutes the
blocks by $\sigma$. The second identity is of the form
\begin{equation}\label{symoperad2}
 \gamma(T; \sigma_1(T_1),\ldots,\sigma_m(T_m)) = \hat\sigma(\gamma (T; T_1,\ldots,T_m)), 
\end{equation}
where $\hat\sigma  \in \Sigma_{n_1+\cdots + n_m}$ is the permutation 
acting on the $i$--th block of $n_i$ indices as $\sigma_i$.

\subsubsection{Algebras over operads}
An algebra $A$ over an operad $\fO$ (in the category of sets) is a set $A$ with an
action of the operad $\fO$ namely operations
\begin{equation}\label{algoper}
\gamma_A: \fO(n) \times A^n \to A  
\end{equation}
satisfying
\begin{equation}\label{gammaAgamma}
\begin{array}{c}
 \gamma_A(\gamma(T;T_1,\ldots, T_n); a_{1,1}, \ldots, a_{1,k_1}, \ldots, a_{n,1}, \ldots, a_{n,k_n}) = \\
 \gamma_A(T; \gamma_A(T_1;  a_{1,1}, \ldots, a_{1,k_1}),\ldots, \gamma_A(T_n; a_{n,1}, \ldots, a_{n,k_n})) \, . 
 \end{array}
 \end{equation}

One interprets here $\gamma_A$ as the operation that takes $n$ inputs $a_i$ in $A$ (an element of $A^n$)
and inserts them as inputs for an $n$-ary operation $T\in \fO(n)$, with the result that the operation then
produces, as output, a new element in the same set $A$. The condition \eqref{algoper} is the usual condition
about actions: one can compose operations in $\fO$ and then apply them to inputs in $A$, or one can
apply the first operations in $\fO$ to inputs in $A$ and then insert the resulting outputs (also in $A$) as
inputs to the second operation in $\fO$ and these two procedures result in the output. 

\subsubsection{Colored operads}
The notion of a {\em colored} operad generalizes the notion of operad recalled above. It is the same
kind of generalization that happens in passing from, say, groups to groupoids, where the product
operation of group elements (which freely applied to any pair of elements) becomes a composition of arrows,
which only applies when the target of the first is the source of the second. In a similar way, in colored operads,
the operad composition operations apply only when there is matching between the colors of the output and 
the input with which it is matched. 

In the case of a colored operad one has a set $\Omega$ of colors (which we take to be a finite set).
A colored operad is a collection $\cO=\{ \cO(c, c_1,\ldots, c_n) \}$ of sets, with $c,c_i\in \Omega$ for $i=1,\ldots,n$,
where the $c_i$ are color labels assigned to inputs of the $n$-ary operations
in $\cO(c, c_1,\ldots, c_n)$, and $c$ is the color label assigned to the output. The 
composition laws then take the form
\begin{equation}\label{colopergamma}
\begin{array}{rl}
\gamma: & \cO(c, c_1,\ldots, c_n) \times \cO(c_1, c_{1,1}, \ldots, c_{1,k_1}) \times \cdots \times
\cO(c_n, c_{n,1}, \ldots, c_{n,k_n})  \\ & \to \cO(c, c_{1,1}, \ldots, c_{1,k_1}, \ldots, c_{n,1}, \ldots, c_{n,k_n}) \, . \end{array}
\end{equation}
They satisfy the same associativity and unit conditions as in the non-colored case, but now with one unit ${\bf 1}_c\in \cO(c,c)$
for each color $c\in \Omega$. In the symmetric case one again as the two equivariance conditions discussed
above, where the action $T\mapsto T\circ \sigma$ maps $\cO(c,c_1,\ldots,c_n)$ to $\cO(c,c_{\sigma(1)},\ldots,c_{\sigma(n)})$
for permutations $\sigma\in S_n$.

An algebra $A$ over a colored operad consists of a collection $A=\{ A_c \}_{c\in \Omega}$ with maps
$$ \gamma_A: \fO(c;c_1,\ldots,c_n) \times A_{c_1}\times \cdots \times A_{c_n} \to A_c $$
with the compatibility \eqref{gammaAgamma}.

Colored operads provide a convenient formalism to describe the generative process of
combinatorial objects with constraints. An example of an application of colored operads
can be found in the theory of formal languages, where a categorical version of the
Chomsky--Sch\"utzenberger representation theorem can be obtained in terms of
the colored operad of spliced words in a category, see \cite{MelZei} and \cite{MelZei2}.
In our setting here, we will use generating systems for colored operads, in the sense
of \cite{Giraudo}, to obtain the constraints on theta positions and theta role matching  
on the freely generated syntactic objects produced by free symmetric Merge, see
section~\ref{ThetaBudSysSec}. 

\subsection{Syntactic objects and the Merge operad}\label{MergeOp}

We recall here the Merge operad discussed in \S 3.6.3.1 of \cite{MCB}. 
This is the operad $\cM$ with $\cM(n)$ the set of all abstract (non-planar)
binary rooted trees with $n$ (non-labelled) leaves. The insertion operations
$T\circ_\ell T'$, for $\ell\in L(T)$ graft the root vertex of $T'\in \cM(m)$ to
the leaf $\ell$ of $T\in \cM(n)$, resulting in a tree $T\circ_\ell T'$ in $\cM(n+m-1)$.

This operad is closely related to what is usually referred to as the
{\em magmatic operad} in the literature, which is the operad freely
generated by a single binary operation. The main difference is that
in the case of the magmatic operad this operation is both non-associative
and non-commutative, and it gives rise to {\em planar} binary
rooted trees as elements of the operad, while here we assume
that the generating binary operation $\fM$ is non-associative and
{\em commutative}, hence we obtain the abstract (non-planar) trees.

\begin{rem}\label{M1}{\rm 
Since this will play an important role in \S \ref{EMIMsemSec}, we also
include in the set $\cM(1)$ of unary operations an operation that we
write as $\fM_{\bullet, 1}$ where $1$ is the unit of the free 
non-associative commutative magma, namely the formal empty tree.
Here $\fM_{\bullet, 1}$ takes as input any $T\in \cM$ (with the root
clued to the $\bullet$ leaf) and outputs $\fM_{\bullet, 1}(T)=\fM(T,1)=T$. Thus, this
is just acting as the identity operation that is the unit ${\bf 1}\in \cM(1)$
of the operad. However, when we include colorings, there will
be a distinction between the unit ${\bf 1}\in \cM(1)$, which will acquire
only a single color, and the operation $\fM_{\bullet, 1}$ that can aquire
two possibly different colors at the input and output. We will return to
this point in \S \ref{EMIMsemSec}. }\end{rem}

As discussed in \S 3.6.3.1 of \cite{MCB}, the set $\cS\cO=\fT_{\cS\cO_0}$ of
syntactic objects, identified with abstract binary rooted trees with leaves
decorated by elements of the set $\cS\cO_0$ of lexical items and 
syntactic features, is an algebra over the Merge operad $\cM$, where
the operad action consists of inserting as inputs the roots of syntactic 
objects $T_\ell \in \fT_{\cS\cO_0}$ at the leaves $\ell\in L(T)$ of an operation $T\in \cM$,
resulting in a new syntactic object $\gamma_\cM(T; \{ T_\ell \})$.

Note that, in principle, in order to apply the operad composition 
and insertions, we need a numbering of the leaves of trees, which
is not given for trees that are non-planar. This is discussed briefly
in \S 3.8.1 of \cite{MCB}. However, we do not need to worry
about this point at all here. In fact, we can proceed by assuming
that the syntactic objects have already been filtered for a head structure.
Thus, trees are in fact pairs $(T,h_T)$ of a tree with an assigned
head function, and as shown in Lemma~3.8.1 of \cite{MCB},
the operad $\cM$ extends as an operad $\cM_h$ consisting of
pairs $(T,h_T)$, since the properties of head function are
compatible with operad composition/insertion. A head function
also determines an ordering of the leaves, determined by the
harmonic head final (or by the harmonic head-initial) planar
embedding determined by the head function. Thus, while
we consider the trees as non-planar, we can index the
operad compositions/insertions using the ordering of
leaves. Thus, we write $T \circ_i T'$ for the insertions
and $\gamma(T; T_1,\ldots, T_n)$ for the compositions,
using this convention on numbering. The resulting
algebra over the operad $\cM_h$ is then given by
${\rm Dom}(h) \subset \fT_{\cS\cO_0}$, syntactic objects
with a head function. The structure of algebra over an
operad is given by the operad action
$\gamma: \cM(n) \times \fT_{\cS\cO_0}^n \to \fT_{\cS\cO_0}$
as illustrated in the simple example
$$ \gamma (\Tree[ $\bullet$ [ $\bullet$ $\bullet$ ]] ; \Tree[ $\alpha$ $\beta$ ], \Tree[ $\gamma$ $\delta$ ], \Tree[ $\epsilon$ [ $\zeta$ $\eta$ ] ] ) = \Tree[ [ $\alpha$ $\beta$ ] [ [ $\gamma$ $\delta$ ] [ $\epsilon$ [ $\zeta$ $\eta$ ] ]  ]] \, . $$

\section{Theta grids, theta criterion and operad bud generating systems} \label{ThetaBudSysSec}

One usually formulates the matching of thematic roles in the following way, see
\cite{ChomskyGB}. There is a finite set $\vartheta$ of thematic relations: agent, experiencer,
theme, goal, recepient, source, location, instrument, etc. Theta-roles are ``bundles of
thematic relations": they are described by a ``thematic grid" or ``theta grid", where to
a predicate (such as V, A, P) that requires a certain number of arguments one assigns
a theta grid with one column per argument, each column representing a different
theta role. One of these columns/theta roles is marked as the unique ``external"
theta role, and all others are marked as ``internal". Theta roles are assigned by
the predicate to its arguments according to the ``theta criterion" rules:
\begin{enumerate}
\item each argument receive one and only one theta role
\item each theta role is assigned to one and only one argument.
\end{enumerate}

\smallskip

In this section we describe how to realize the assignment of theta roles in terms of 
colored operads and generating systems for colored operads, and how the theta
criterion is implemented in this formulation.

\subsection{Bud generating systems}\label{BudSec}

We recall from \cite{Giraudo} the combinatorial generating systems for colored
operads and formal languages associated to colored operads. 

Given an ordinary operad $\fO$ and a finite set $\Omega$, the bud-operad
$\bB_\Omega(\fO)$ is the colored operad with $\Omega$ as the set of colors
and with 
\begin{equation}\label{BudO}
\bB_\Omega(\fO)(n) := \Omega \times \fO(n) \times \Omega^n \, ,
\end{equation}
namely all the possible assignments of colors in $\Omega$ to the inputs and the output of
the operations in $\fO(n)$. The operad composition, described in terms of insertion
operations is given by
$$ (c,T, c_1,\ldots, c_n) \circ_i (c', T', c'_1,\ldots, c'_m)= (c, T\circ_{\fO,i} T', c_1, \ldots, c_{i-1}, c'_1, \ldots, c'_m, c_{i+1}, \ldots, c_n) \ \ \ \text{ if } c_i=c' \, , $$
and is undefined if $c_i \neq c'$. The pruning map is the morphism of operads $\wp: \bB_\Omega(\fO) \to \fO$
that forgets the colors. 

\begin{defn}\label{budgensef} {\rm \cite{Giraudo}.}
A bud generating system $\bB=(\fO,\Omega,\cR, \cI,\cT)$ consists of an operad $\fO$,  a finite set of colors $\Omega$,
a finite subset $\cR  \subset \bB_\Omega(\fO)$ of operations in the bud-operad \eqref{BudO} (the set of coloring rules), 
a set $\cI$ of initial colors and a set $\cT$ of terminal colors. 
\end{defn}

\begin{defn}\label{budderive} {\rm \cite{Giraudo}.}
An operation $x=(c,T,\underline{c})$ in $\bB_\Omega(\fO)$ is derivable in $\bB$ from another $y=(c',T',\underline{c'})$ if
there are $r_1,\ldots, r_N\in \cR$ such that $y$ is obtained from $x$ through a sequence of compositions $\circ_{i_k} r_k$.
One writes $y \to_{\bB} x$, for $x$ derivable from $y$ in $\bB$. 
The language $\cL(\bB)$ is the set of all $x\in \bB_\Omega(\fO)$ that are derivable in $\bB$ from a unit ${\bf 1}_c$,
with $c\in \cI$, and $x$ has all inputs colored by colors in $\cT$,
\begin{equation}\label{LBlang}
 \cL(\bB)=\{ x=(c,T,\underline{c})\in \bB_\Omega(\fO)\,|\, {\bf 1}_c \to_{\bB} x, \, \, c\in \cI, \, \, \underline{c}\in \cT^n \} \, . 
\end{equation} 
\end{defn}

For our setting, it suffices to restrict to the case where the initial set of colors consists of
all the nonterminal colors, $\cI =\Omega \smallsetminus \cT$. We then have the following result, which
is a direct consequence of the definitions above. 

\begin{lem}\label{opBOalgLB}
A bud generating system $\bB=(\fO,\Omega,\cR, \Omega \smallsetminus \cT,\cT)$ with 
initial colors $\cI =\Omega \smallsetminus \cT$ determines a colored operad $\fO_{\Omega, \bB}$ with
\begin{equation}\label{opBO}
\fO_{\Omega,\bB}(n):= \{ x=(c,T,\underline{c})\in 
\bB_\Omega(\fO)(n) \,|\, {\bf 1}_c \to_{\bB} x, \, \, c\in \Omega \smallsetminus \cT, \, \, 
\underline{c}\in (\Omega \smallsetminus \cT)^n \} \, .
\end{equation}
The language $\cL(\bB)$ of \eqref{LBlang} is an algebra over the colored operad $\fO_{\Omega,\bB}$. 
Let $\wp_{in}: \bB_\Omega(\fO) \to \sqcup_n \fO(n)\times \Omega^n$ be the projection
$\wp_{in}(c,T,\underline{c})=(T,\underline{c})$ that forgets the output coloring, and let
$\wp_\cT: \cL(\bB) \to \sqcup_n \fO(n)\times \Omega^n$ be the restriction of $\wp_{in}$ to $\cL(\bB)$.
The image $\wp_\cT(\cL(\bB))$ is an algebra over the operad $\fO$. 
\end{lem}

One should think of a colored operad $\fO_{\Omega,\bB}$ obtained in this way as a colored
version of the operad $\fO$, where the coloring happens according to local coloring
rules specified by the set of generators $\cR$, consistently propagated via operad composition.
The language $\cL(\bB)$, seen as an algebra over the operad $\fO_{\Omega,\bB}$,
represents the labelled structures obtained from the generating system, with the
action of the colored operad. In our main application here, the language $\cL(\bB)$
will provide a colored version of the set of syntactic objects $\fT_{\cS\cO_0}$, with 
the structure of algebra over an operad $\fO_{\Omega,\bB}$ of colored non-planar 
binary rooted trees that is a colored version of $\cM_h$, where the local coloring moves 
represent the basic theta-grids. 

\subsection{The bare theta bud system}\label{BareThetaBudSec}

We present two bud systems over the same Merge operad $\cM$, corresponding to two different
set of colors, one that only contains the theta roles, and one that also has a marker for 
non-theta positions. The relation between these two levels of structure will account for the 
consistency checking on the coloring that corresponds to the linguistic theta criterion, while
the bud system that includes non-theta positions can more directly relate to the argument
structure and the syntactic head and complement structure. We will refer to the setting that
only considers theta roles as the ``bare theta bud system" and to the one that also allows for
non-theta positions as simply the ``theta bud system" (or the ``complete theta bud system").

We first construct an appropriate set of colors $\Theta$ for describing theta theory in
terms of colored operads. 
Let $\vartheta=\{ \theta_i \}_{i=1}^N$ denote the finite set of all
thematic roles (which we write as $\theta_i$), endowed with a preorder structure. 
These roles include agent, experiencer, theme, 
recepient, location, instrument, etc., and is understood to include 
{\em all} the usual theta roles. The preorder structure accounts for theta hierarchies.

Note that we consider on $\vartheta$ a preorder rather than a partial order. 
A partial order satisfies reflexivity, antisymmetry, and transitivity, while a preorder
is only required to satisfy reflexivity and transitivity. The antisymmetry relation ($x\leq y$
and $y\leq x$ $\Rightarrow$ $x=y$) appears too strong in the context of theta hierarchies. 

We also introduce a label $\uparrow, \downarrow$ and write the thematic roles
as either $\theta_i^\uparrow$ or $\theta_i^\downarrow$, to distinguish a {\em giver} 
$\theta_i^\uparrow$ of theta role $\theta_i$ from a {\em receiver}  $\theta_i^\downarrow$
of the same theta role. We then extend the set $\vartheta$ defined above to  
\begin{equation}\label{varthetaroles}
 \vartheta=\{ \theta_i^\uparrow \}_{i=1}^N\sqcup \{ \theta_i^\downarrow \}_{i=1}^N \, . 
\end{equation} 
The preorder structure extends just by forgetting the $\uparrow, \downarrow$ labels.

\begin{defn}\label{barethetacolors}
Consider the Cartesian product sets $\vartheta^n$, for $1\leq n \leq n_{\rm max}$
where $n_{\rm max}$ is the largest valency of a predicate (the largest number of different theta roles
that a predicate can assign), with $\vartheta$ as in \eqref{varthetaroles}. Let
$$ \Theta_b :=\sqcup_{n=1}^{n_{\rm max}} \,\, \vartheta^n \, . $$
We then define the color set as
\begin{equation}\label{ThetaColor}
\Theta = \Theta_b \sqcup (\cS\cO_0\times \Theta_b) \, ,
\end{equation}
where we take
\begin{equation}\label{terminitialcol}
 \cT_{\Theta,b} = \cS\cO_0\times \Theta_b \ \ \ \text{ and } \ \ \  \cI_{\Theta,b}=\Theta\smallsetminus \cT_{\Theta,b}=\Theta_b  
 \end{equation}
to be the sets of terminal and initial colors, respectively.
A theta-hierarchy is a vector $\underline{\theta}=(\theta_1,\ldots,\theta_n) \in \vartheta^n$ such that its entries are ordered with
respect to the partial order structure on the set $\vartheta$, namely $\theta_1 > \theta_2 > \ldots > \theta_n$. (Since the
preorder relation does not depend on the labels $\uparrow,\downarrow$ we do not write them explicitly here.)
\end{defn}

\begin{rem}\label{gridrem}{\rm 
Note that the notion of valence that we consider here may differ slightly from the one adopted in some
of the linguistics literature, as here we describe a theta grid $\underline{\theta}=(\theta_1,\ldots,\theta_n)$
as having valence $n$, while when one decomposes it in the form $\underline{\theta}=(\theta_E, \underline{\theta}_I)$
of an external and internal part, it is sometimes common to assign it valence $n-1$ (the size of the internal
$\underline{\theta}_I$), so that for example a predicate like ``sleep" is assigned valence $0$ rather than $1$
and ``give" is assigned valence $2$ rather than $3$. We adopt here the convention where the external $\theta_E$
is also counted as part of the valence, so ``sleep" is unary and ``give"  is ternary. }
\end{rem}

\begin{defn}\label{inouttheta}
An input theta-color grid $\underline{c}=(c_0,c_1,\ldots, c_{n-1}, c_n)$ is an element in $\Theta_b^{n+1}$ of
the form 
\begin{equation}\label{cgrid}
 \underline{c}=(\theta_E^\downarrow, \theta_{I,1}^\downarrow, \theta_{I,2}^\downarrow, \ldots , 
\theta_{I,n-1}^\downarrow, \underline{\theta}^\uparrow) 
\end{equation}
where $\underline{\theta}=(\theta_E, \underline{\theta}_I)$ and 
$\underline{\theta}_I=(\theta_{I,1}, \ldots, \theta_{I,n-1})$ 
is a theta hierarchy, $\theta_{I,1}>\theta_{I,2}> \cdots > \theta_{I,n-1}$.  This includes also the possibility of just $\underline{\theta}=\theta_E$ with no $\underline{\theta}_I$.
\end{defn}

For example, in a sentence like {\em Brian gave the book to Mary}, the verb {\em to give} is the predicate
that assigns three theta roles: $\theta_E=$agent (Brian), $\theta_{I,1}=$theme (the book), and $\theta_{I,2}=$goal (to Mary).
In this case the theta grid is $\underline{\theta}=(\text{agent}_E, \text{theme}_{I,1}, \text{goal}_{I,2})$ and the associated
input theta-color grid is of the form
\begin{equation}\label{givegrid}
 \underline{c}=((\text{agent}_E^\downarrow, \text{theme}_{I,1}^\downarrow, \text{goal}_{I,2}^\downarrow), \underline{\theta}^\uparrow) 
\end{equation} 
which means that in the structure that corresponds to the sentence {\em Brian gave the book to Mary} there
are four positions: one (the predicate) that assigns all three theta roles, and will be contributing the component
$\underline{\theta}^\uparrow$ of the theta-color grid, and three other separate positions, each of which
receives one of these three theta roles. These three other positions contribute the terms $(\text{agent}_E^\downarrow, 
\text{theme}_{I,1}^\downarrow, \text{goal}_{I,2}^\downarrow)$ of the theta-color grid \eqref{givegrid}. 

\begin{ex}\label{giveEx}{\rm 
Note that the ordering of the components of $\underline{c}$ in the input theta-coloring grid do not
correspond here to the ordering in which the theta giver and receivers would appear in a sentence.
For example, in the sentence {\em Brian gave the book to Mary} the order appears to be
$(\text{agent}_E^\downarrow, \underline{\theta}^\uparrow, \text{theme}_{I,1}^\downarrow, \text{goal}_{I,2}^\downarrow)$,
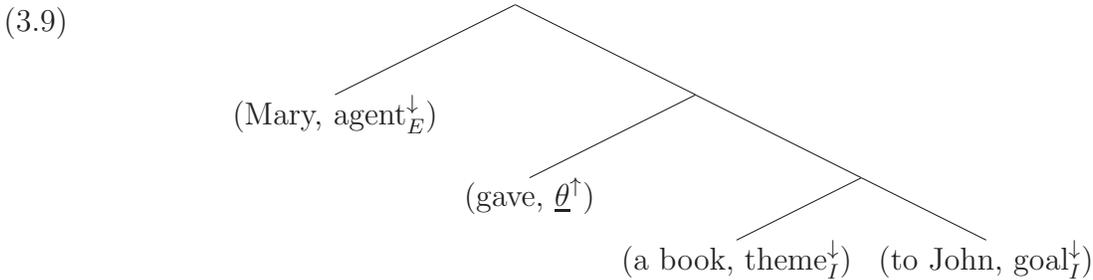
\begin{equation}\label{gave1}
\Tree[ {(Mary, $\text{agent}_E^\downarrow$)} [ {(gave, $\underline{\theta}^\uparrow$)} [ {(a book, $\text{theme}_I^\downarrow$)}  {(to John, $\text{goal}_I^\downarrow$)} ] ] ]  
\end{equation}
rather than the order of $\underline{c}$ in \eqref{givegrid} above. 
We will discuss below how these reorderings are
obtained as a result of movement, and what the distribution of theta roles
for the syntactic objects corresponding to this and other similar examples 
look like in the model developed here. }
\end{ex}

We first introduce a simple bud generating system that accounts for syntactic
objects with theta roles satisfying the theta condition. We then extend it to a
more realistic model that also allows for the presence of non-theta positions
and the effect of movement.

\begin{defn}\label{barethetabud}
The bare theta bud generating system is the collection 
$$ \bB_{\Theta,b}=(\cM_h, \Theta_b, \cR_b, \cI_{\Theta,b}, \cT_{\Theta,b}), $$
with $\cM_h$ the Merge operad with head structure, $\Theta_b$, $\cI_{\Theta,b}$, and $\cT_{\Theta,b}$ as above, and with
the set $\cR_b \subset \bB_{\Theta,b}(\cM_h)$ given by elements $(c,T,\underline{c})$ satisfying the following properties:
\begin{enumerate}

\item The tree $T$ is a comb tree with coloring given by an arbitrary $c\in \cI_{\Theta,b}$ at the root and an input
theta-color grid $\underline{c}$ (as in Definition~\ref{inouttheta}) at the leaves, 
\begin{equation}\label{barethetatree}
 (c,T,\underline{c}) = \Tree[.$c$ $\theta_E^\downarrow$ [.$\theta_E^\uparrow$ $\theta_{I,1}^\downarrow$ 
[.$\underline{\theta}_1^\uparrow$ $\theta_{I,2}^\downarrow$ [.$\ldots$ $\ldots$  [ .$\underline{\theta}_{n-2}^\uparrow$ 
$\theta_{I,n-1}^\downarrow$ [.$\underline{\theta}_{n-1}^\uparrow$ $\theta_{I,n}^\downarrow$ $\underline{\theta}^\uparrow$ ] ]]]] ] 
\end{equation}
where  
\begin{equation}\label{thetakup}
\underline{\theta}_k^\uparrow:=(\theta_E^\uparrow, \theta_{I,1}^\uparrow,\ldots,\theta_{I,k}^\uparrow). 
\end{equation} 
This includes also the possibility with $\underline{\theta}=\theta_E$ with no $\underline{\theta_I}$ and 
$$ (c,T,\underline{c}) = \Tree[.$c$ $\theta_E^\downarrow$ $\underline{\theta}^\uparrow$ ] $$
\item The leaf $\ell\in L(T)$ decorated by $\underline{\theta}^\uparrow$ is the head of the tree $\ell=h_T(v)$, with
$v$ the root vertex of $T$. 
\end{enumerate}
In addition to these elements, $\cR_b$ also contains elements $(c,T,\underline{c})$ of the same
form but where some (or all) of the leaves are decorated by terminals $(\alpha_0, \theta_E^\downarrow)$, $(\alpha_1, \theta_{I,1}^\downarrow)$, $\ldots$, $(\alpha_{n-1}, \theta_{I,n-1}^\downarrow)$, $(\alpha_n,\underline{\theta}^\uparrow)$.
We refer to the resulting language $\cL(\bB_{\Theta,b})$ as the language of bare theta structures.
\end{defn}

Here the pair $(\alpha_n,\underline{\theta}^\uparrow)$ 
consists of a lexical item and a syntactic feature (eg V,A,P) indicating a predicate that has a grid
$\underline{\theta}$ of theta roles to assign, and the $(\alpha_1, \theta_E^\downarrow)$ and $(\alpha_i,\theta_{I,i}^\downarrow)$
are lexical items with syntactic features (e.g., N) that can receive a theta role $\theta_E$ or $\theta_{I,i}$.

\begin{rem}\label{colorVE}{\rm
Here we always assume that, in a binary rooted tree $T$, if a non-root vertex $v$ is labelled by a color $c$ then
the unique edge above $v$ (in the direction of the root) is also colored by the color $c$, so we can speak
equivalently of the colors as either vertex of edge colors, following this convention. 
}\end{rem}

The generators $(c,T,\underline{c})$ of the form \eqref{barethetatree} should be interpreted as a comb tree where
a certain theta grid $\underline{\theta}$ is inserted at the bottom, at the leaf marked by $\underline{\theta}^\uparrow$.
This behaves like physical momenta in Feynman graphs: it propagates through the tree satisfying a momentum conservation
rule at each node that it traverses, so that at each node on the path to the root, a part of the theta grid is diverted down the other edge
and received by the leaf marked $\theta_{I,i}^\downarrow$, while the remaining part $\underline{\theta}_k^\uparrow$
as in \eqref{thetakup} continues to flow up the tree, until at the last node $\theta_E^\downarrow$ is
also received by the upmost leaf. Since there are no more theta roles to assign, the root vertex remains
free to either receive or assign another independent theta role $c$ in any larger structure in which it will 
participate. 

The coloring rules $(c,T,\underline{c})\in \cR_b$ are templates for theta relations satisfying the
theta criterion, for all different possible valencies. We phrase the theta criterion in the following way. 

\begin{defn}\label{thetacrit}
An element $(c,T,\underline{c})\in \bB_\Omega(\fO)(n)$ satisfies the theta criterion if
\begin{enumerate}
\item the leaf that is the head of $T$ has a color $c_T\in \vartheta^k$, for $k=\# L(T)-1$, containing 
only $\uparrow$ labels, while all other leaves have labels $c_i=\theta_i^\downarrow \in \vartheta$;
\item any $\theta_i^\downarrow \in \vartheta$
occurring as a component of $\underline{c}$ is matched by a component $\theta_i^\uparrow$ occurring in the tuple 
$\underline{\theta}^\uparrow$ at the head leaf.
\end{enumerate}
\end{defn}

This definition clearly corresponds to the usual phrasing of the theta criterion, 
in the sense that the vector $\underline{c}$ realizes a complete matching to the 
theta roles assigned in the grid $\underline{\theta}^\uparrow$ at the
head $\ell=h_T(v)$ with the arguments (the other leaf), where each theta role $\theta_i^\uparrow$ occurring in 
$\underline{\theta}^\uparrow$ is uniquely assigned to the argument carrying the $\theta_i^\downarrow$ label
and each argument receives exactly one theta role, since each other leaf has a label in $\vartheta$ and not in
any $\vartheta^k$ with $k>1$.

Note that the color $c$ assigned at the root vertex does {\em not} participate in the theta criterion matching. Indeed,
as mentioned above, the color $c$ at the root represents a theta role that the entire structure $T$ 
either assigns or receives when it becomes part of a larger structure. This allows for cases where
theta roles are received or assigned by an entire phrase rather than by a single lexical item. Thus,
the same structure $T$, with the same 
matching set of theta roles $\underline{c}=(\theta_E^\downarrow,\underline{\theta}_I^\downarrow,\underline{\theta}^\uparrow)$ at the leaves
can occur in $\bB_\Omega(\fO)$ with different root colors $c$, which correspond to different 
possible theta roles that the entire structure $T$ can play inside a larger structure: 
each of these possibilities corresponds to a different element of $\bB_\Omega(\fO)$.

\begin{lem}\label{thetacrit}
The bare theta structures $(c,T,\underline{c}) \in \cL(\bB_{\Theta,b})$ satisfy the theta criterion.
\end{lem}

\proof
By construction the generators \eqref{barethetatree} have balanced theta roles assigned
at the leaves so every component of the insertion $\underline{\theta}^\uparrow$ is matched by
a component $\theta_i^\downarrow$ at one of the other leaves. 
All the bare theta structures in $\cL(\bB_{\Theta,b})$ are then 
constructed from these generators through color-matching operad insertions, in such a way
that, in the resulting object, all the leaves carry terminal colors. Each insertion corresponds to
matching a theta role given or received by a leaf to the same theta role at the root
of the inserted tree, to signify that, at the vertex where the operad insertion happens, the 
theta role is assigned or received by the entire phrase below the insertion point. 
Since every coloring rule $(c,T,\underline{c})\in \cR_b$ has $\underline{c}$ satisfying 
the theta criterion as in Definition~\ref{thetacrit}. 
Note that the labeling $c$ at the root of $T$ is not part of the
matching of theta roles for $T$, as it will become part of the matching of
theta roles of another structure $T'$ under the colored operad insertions $T'\circ_i T$. 
Since theta roles (at the leaves) are matched according to the theta criterion in each
$(c,T,\underline{c})\in \cR_b$ they remain matched according to the theta criterion
in any colored operad insertion and hence in every element of the language 
$\cL(\bB_{\Theta,b})$. Thus all the bare theta structures satisfy the theta criterion.
\endproof

Note, however, that not all the $(c,T,\underline{c})\in \bB_\Omega(\fO)(n)$ that satisfy
the theta criterion as stated in Definition~\ref{thetacrit} are in $\cL(\bB_{\Theta,b})$.
For example, cases such as
\begin{equation}\label{thetaother}
 \Tree[ $\theta_E^\downarrow$ [ $\underline{\theta}^\uparrow$ [ $\theta_{I,1}^\downarrow$ $\theta_{I,2}^\downarrow$ ] ] ]  
\end{equation} 
are not directly obtainable from operad insertions starting from generators of the form \eqref{barethetatree},
hence they are not in $\cL(\bB_{\Theta,b})$. The language $\cL(\bB_{\Theta,b})$ accounts for a very restricted
class of syntactic objects satisfying the theta criterion. We will see below why they in fact suffice to account
for theta theory correctly, once non-theta positions are also included. 

The insertion of non-theta positions is necessary: indeed, 
the bare theta structures clearly do not suffice to model the assignment of theta
roles to syntactic objects, both because they do not capture all the configurations
of theta role assignments that we want to represents (for instance cases like Example~\ref{giveEx})
and also because typically there are several additional parts of
a syntactic object that do not participate in the mandatory matching of theta roles, since they are not
part of the argument structure of predicates (the parts that have mandatory
relations). The additional structures typically include adjuncts, which are not
mandatory relations, and also positions associated to movement that are
not receivers or givers of theta roles. Examples include ``free adjunct constructions" such
as {\em Walking home he found a dollar}, in sentence initial position, or {\em He found a dollar walking home},
sentence finally, or the ``nominative absolute constructions" such as {\em His father being a sailor, John
knows all about boats}, or the analogous sentence final version, or also cases like the ``augmented
absolute construction" such as {\em With the children asleep, John watched TV}, which again can
also occur in sentence final position. 
Since there can be, in principle, arbitrarily long chains 
of adjuncts that can be added, a set of coloring rules that realistically describes the 
assignments of theta roles should allow for these insertions of an arbitrary number of elements
that are not givers or receivers of theta roles. We show in \S \ref{ThetaBudSec}
how this can be done, while still using a relation to the bare theta structures of
$\cL(\bB_{\Theta,b})$ to ensure that the theta criterion is satisfied.

\subsection{Theta bud system and the colored operad of theta relations}\label{ThetaBudSec}

We introduce a larger set of colors $\Theta \supset \Theta_b$ and a refinement of the
coloring rules $\cR_b$ of the bare theta bud system of Definition~\ref{barethetabud}, which
allows for the presence of non-theta positions. 

\begin{defn}\label{Theta0}
Consider the same set $\vartheta=\{ \theta_i^\uparrow \}_{i=1}^N\sqcup \{ \theta_i^\downarrow \}_{i=1}^N$ 
as in \S \ref{BareThetaBudSec},
and the corresponding $\vartheta^n$, for $1\leq n \leq n_{\rm max}$. 
Also write $\vartheta^0:=\{ \theta_0 \}$ where the element $\theta_0$
does not mark a theta-role, but the absence of any, namely a marker for a non-theta position. Let
$$ \Theta_0:=\sqcup_{n=0}^{n_{\rm max}} \,\, \vartheta^n \, . $$
As in the case of the bare system in \S \ref{BareThetaBudSec},
define the color set as
\begin{equation}\label{ThetaColor}
\Theta = \Theta_0 \sqcup (\cS\cO_0\times \Theta_0) \sqcup \{ (1,\theta_0) \} \, ,
\end{equation}
where the element $(1, \theta_0)$ is the pair of the formal empty tree $1$ (the unit
of the magma $\fT_{\cS\cO_0}$) marked with the non-theta marker. 
Take as the subset of terminal colors
$$ \cT_\Theta = (\cS\cO_0\times \Theta_0) \sqcup \{ (1,\theta_0) \} \, , $$
with $\cI_\Theta=\Theta\smallsetminus \cT_\Theta=\Theta_0$.
\end{defn}

Note here that non-theta positions are neither givers nor receivers, so they are
only marked by $\theta_0$ without $\uparrow, \downarrow$. Moreover, while in
the previous case where all markings where by theta roles, and where we
only considered non-trivial elements in the magma $\fT_{\cS\cO_0}$, here we
can also consider the unit of the magma (the formal empty tree) with attached
label $\theta_0$ (as obviously the empty tree does not participate in any theta
relations). This seemingly insignificant addition of the terminal color $(1,\theta_0)$ 
will in fact play an important role in the compatibility of theta role and theta
criterion with movement, and with the dichotomy in semantics between 
External Merge and Internal Merge, as we will discuss below.

\begin{defn}\label{ThetaBudSys}
The theta bud generating system (or complete theta bud generating system) consists of
data $\bB_\Theta=(\cM_h, \Theta, \cR_\Theta,  \cI_\Theta, \cT_\Theta)$ with $\Theta, \cI_\Theta, \cT_\Theta$
as in Definition~\ref{Theta0} and with $\cM_h$ the Merge operad with head structure, as 
recalled in \S \ref{MergeOp}. The set $\cR_\Theta$ of coloring rules then consists 
of elements $(c,T,\underline{c})\in  \bB_\Theta(\cM_h)$ that belong to one of the following cases:
\begin{enumerate}
\item For $\underline{\theta}=(\theta_E, \underline{\theta}_I)$ with $\underline{\theta}_I=(\theta_{I,1}, \ldots, \theta_{I,n-1})$ 
a theta hierarchy, $\theta_{I,1}>\theta_{I,2}> \cdots > \theta_{I,n-1}$, and for an arbitrary $c\in \cI_\Theta$, elements
of the forms \eqref{Rule1}, \eqref{Rule2}, \eqref{Rule3}, \eqref{Rule3b} are in $\cR_\Theta$:
\begin{equation}\label{Rule1}
 (c,T,\underline{c}) =  \Tree[ .$c$ [ $\theta_E^\downarrow$ $\theta_E^\uparrow$ ] ]
\end{equation} 
\begin{equation}\label{Rule2}
 (c,T,\underline{c}) = \Tree[ .$\theta_E^\uparrow$ [ $\theta_{I,1}^\downarrow$ {$(\theta_E^\uparrow,  \theta_{I,1}^\uparrow)$} ] ] 
\end{equation} 
\begin{equation}\label{Rule3}
 (c,T,\underline{c}) =  \Tree[ .$\underline{\theta}_k^\uparrow$ [ {$\theta_{I,k+1}^\downarrow$}  {$\underline{\theta}_{k+1}^\uparrow$} ] ]  
\end{equation} 
\begin{equation}\label{Rule3b}
 (c,T,\underline{c}) = \Tree[ .$\underline{\theta}_{n-1}^\uparrow$ [ $\theta_{I,n}^\downarrow$ $\underline{\theta}^\uparrow$ ] ] 
\end{equation} 
The leaf marked with $\underline{\theta}^\uparrow$ in \eqref{Rule1} and \eqref{Rule3b} is the head of the tree.
\item The same cases \eqref{Rule1}, \eqref{Rule2}, \eqref{Rule3}, \eqref{Rule3b}, where one or more of the leaves 
have terminal symbols, respectively $(\alpha_0, \theta_E^\downarrow)$, $(\alpha_i, \theta_{I,i}^\downarrow)$,
$(\alpha_n, \underline{\theta}^\uparrow)$, 
where $\alpha_i \in \cS\cO_0$ are lexical items and syntactic features that are givers and receivers
of the corresponding theta roles. 
\item For any $c\in \cI_\Theta$, an element of the form
\begin{equation}\label{Rule4}
 (c,T,\underline{c}) =\Tree[.$c$ [ $c$ $\theta_0$ ] ] 
\end{equation} 
\item For any $c' \in \cI_\Theta$, elements of the form
\begin{equation}\label{Rule5}
 \Tree[.$\theta_0$ [ $c'$ $(1,\theta_0)$ ] ]   \ \ \ \text{ and } \ \ \   \Tree[.$\theta_0$ [ $(\alpha,c')$  $(1,\theta_0)$ ] ] 
\end{equation} 
for either a non-terminal $c'$ or a terminal $(\alpha, c')$. 
\end{enumerate}
The elements of the language $\cL(\bB_\Theta)$ are the ``unbalanced theta structures".
\end{defn}

The reason why we refer to the elements of $\cL(\bB_\Theta)$ as ``unbalanced" theta structures
is because, unlike the elements of $\cL(\bB_{\Theta,b})$, they do not automatically satisfy the
theta criterion. 

Here the leaf marked with $\underline{\theta}^\uparrow$ is the head of the tree and, if marked by a terminal, the
corresponding lexical item and feature $\alpha_n$ is a predicate that assigns a theta grid $\underline{\theta}$.

The structures generated by the bud generating system of Definition~\ref{ThetaBudSys} are
much less constrained and much more general than those in $\cL(\bB_{\Theta,b})$
obtained from generators of the form \eqref{barethetatree}. For example, we obtain
(non-planar) trees that look like  
\begin{equation}\label{thetatheta0tree}
\Tree[ .$c$ [ $\theta_E^\downarrow$ $\theta_0$ ] [ $\theta_0$  [ $\ldots$ [ $\theta_0$ [ [ $\theta_{I,n-1}^\downarrow$ $\theta_0$ ] [ $\theta_0$ [ [ $\theta_0$ $\theta_{I, n}^\downarrow$ ] $\underline{\theta}^\uparrow$ ]]]]]]] \, . 
\end{equation}
We will show in \S \ref{MoveTheta} that the structures obtained in this way also account for structures obtained
via movement from those in $\cL(\bB_{\Theta,b})$, including those occurring in cases like
Example~\ref{giveEx}, and their compatibility with the grid of theta roles
and the theta condition.

\subsection{Balanced theta structures and the theta criterion} \label{BalTheta}

We then obtain the following relation between the bare theta structures, these unbalanced
theta structures and the balanced complete theta structures that will account for the correct assignment
of theta roles to syntactic objects, including the non-theta positions.

 \begin{thm}\label{ThetaThetab}
 The elements of $\cR_b$ are derivable in $\cB_\Theta$ from elements of $\cR_\Theta$, and
 $\cL(\bB_{\Theta,b}) \subset \cL(\bB_\Theta)$ is the subset of all elements of $\cL(\bB_\Theta)$ that
 contain no non-theta positions and satisfy the theta criterion. It is also characterize as the
 set of $(c,T,\underline{c})\in \cL(\bB_\Theta)$ where only generators of cases (1) and (2) of 
 Definition~\ref{ThetaBudSys} are used and each non-leaf vertex $v_\ell$ of $T$ that is the 
 projection of an $\ell\in L(T)$ under the head function is the output of a generator of the form \eqref{Rule1}.
 \end{thm}
 
 \proof
 The four elements of $\cR_\Theta$ listed in cases (1) and (2) of Definition~\ref{ThetaBudSys}
 suffice to obtain via operad insertions all the elements of $\cR_b$. Elements of 
 $\cL(\bB_\Theta)$ that contain no non-theta positions may or may not satisfy the theta criterion, but
 they do have the property that, at every non-leaf vertex $v$, a ``conservation law" for the theta roles is satisfied,
 where one of the two vertices below $v$ (the head direction) has a tuple of theta roles with $\uparrow$, the
 other vertex below $v$ has one of these theta roles with $\downarrow$ and $v$ has the remaining
 theta roles with $\uparrow$. This is the case since each of the 
 generators (all of which have a root vertex and two leaves) has 
 a theta role of the form $\theta_i^\uparrow$ present (among others) at the root leaf and a 
 theta role $\theta_i^\downarrow$ present (by itself) at the other leaf, with the remaining theta roles
 with $\uparrow$ at the root vertex. This local rule then remains valid after all the operad
 compositions are performed. This rule means that there is a local form of the theta
 criterion that is satisfied and the question is whether it is also satisfied globally over the resulting
 syntactic object. This happens if and only if at each vertex $v_\ell$ of $T$ (a maximal projection of
 a head $\ell \in L(T)$ that projects) the two vertices below $v_\ell$ carry a matching pair $\theta_i^\uparrow$
 and $\theta_i^\downarrow$, i.e., the subtree with these tree vertices is a generator of the form \eqref{Rule1}.
 Indeed, this condition ensures that each chain of compositions of generators that starts with the insertion
 of an $\underline{\theta}^\uparrow$ at the head leaf $\ell$ traverses the right number of nodes along the
 path $\gamma_\ell$ from the leaf $\ell$ to its maximal projection $v_\ell$ to match every $\theta_i^\uparrow$ in
 $\underline{\theta}^\uparrow$ with a $\theta_i^\downarrow$, so that the last matching of $\theta_E^\uparrow$
 and $\theta_E^\downarrow$ happens at $v_\ell$, which then acquires a new color $c$ that participates
 in the next matching of theta roles at the next maximal projection higher up in the tree. 
  \endproof

  \subsubsection{Coproducts}\label{coprodSec}
  We recall here, from Chapter~1 of \cite{MCB}, the coproduct structure on workspaces, which
  is the main algebraic operation underlying both Merge and parsing at the syntax-semantics
  interface. Workspaces are disjoint unions of syntactic objects, that is, binary forests with leaves
  labelled by elements in $\cS\cO_0$. We denote the set of workspaces by $\fF_{\cS\cO_0}$. The
  set of syntactic objects $\fT_{\cS\cO_0}$ is the subset of $\fF_{\cS\cO_0}$ that consists of
  forests with a single connected component (that is, trees). We also denote by $\cV(\fF_{\cS\cO_0})$
  the vector space (say, over $\Q$) spanned by the workspaces, namely the set of formal linear
  combinations of workspaces. The vector space $\cV(\fF_{\cS\cO_0})$ has additional algebraic
  structure, namely a grading (by number of leaves of the forests) and a product and coproduct that make it
  a Hopf algebra. The product is just the disjoint union $\sqcup$ of forests, but the coproduct
  is the interesting part that allows for the extraction of accessible terms for movement.
  To be more precise, there is more than one form of the coproduct, and it
  will be important here to distinguish them. The algebraic properties vary slightly in these
  different cases, but for simplicity, here as in \cite{MCB}, we will just use the term ``Hopf algebra
  of workspaces" to denote this whole structure with all its variants. What we need to recall here is the
  form of the coproduct, which can be written as
  \begin{equation}\label{DeltaAcc}
   \Delta(T)=\sum_{\underline{v}=\{ v_1, \ldots, v_k\}} F_{\underline{v}} \otimes T/F_{\underline{v}}\, , 
  \end{equation} 
  where the sum is over disjoint collections of accessible terms $T_{v_i}\subset T$ with
  $F_{\underline{v}}=T_{v_1}\sqcup \cdots \sqcup T_{v_k}$. The sum lists all the possibilities
  of extractions of accessible terms, starting with the cases with no extraction or extraction of
  the entire tree (forming the primitive part of the coproduct $T\otimes 1 + 1 \otimes T$),
  followed by the cases of extraction of a single accessible term $T_v \otimes T/T_v$, then
  multiple extractions. Another way of writing the terms of this coproduct is in terms of {\em admissible cuts}
  \begin{equation}\label{Deltacut}
   \Delta(T)=\sum_{\underline{v}} F_{\underline{v}}  \otimes T/ F_{\underline{v}} = \sum_C \pi_C(T) \otimes \rho_C(T)\, , 
  \end{equation} 
  where an admissible cut $C$ is a cut of a number of edges in the tree with the property that no two of them are
  on the same path from the root to a leaf. The equivalence between extraction of accessible terms and admissible
  cuts comes from the fact that any edge $e$ cut by $C$ produces the extraction of the accessible term $T_v$ rooted
  at the vertex $v$ at the lower end of $e$ (away from the root). The term $\pi_C(T) =F_{\underline{v}}$ is the forest
  that falls off the tree as a result of the cut $C$ and $\rho_C(T)=T/ F_{\underline{v}}$ is what remains attached to the root.
  Again this sum over admissible cuts includes the trivial cases where $C$ cuts nothing or everything 
  resulting in the two terms in the primitive
  part $1 \otimes T + T \otimes 1$ of the coproduct, while the remaining terms are the nontrivial cuts.
  
  \subsubsection{Recursive checking of theta criterion}
  The forms of the coproduct differ in the description of the remainder
  term $\rho_C(T)=T/ F_{\underline{v}}$ after the cut is performed. We are interested here in two
  possibilities. The ``deletion" quotient $T/^d F_{\underline{v}}$ is obtained by removing the edges in the cut $C$
  and then taking the maximal full binary tree that is obtained from what remains 
  after performing some edge contractions to eliminate non-branching vertices. The ``contraction" quotient
  $T/^c F_{\underline{v}}$ is obtained by shrinking each $T_{v_i}$ to its root vertex $v_i$ in $T$, which becomes
  a new leaf, labelled by the syntactic features of $T_{v_i}$.  As mentioned these two forms of the coproduct
  have different algebraic properties (we refer the reader to \cite{MCB} for a more detailed discussion). From the
  linguistic perspective, in the extraction of accessible terms for movement, the quotient $T/^c F_{\underline{v}}$
  maintains the trace of movement (needed for interpretation at the CI interface), while $T/^d F_{\underline{v}}$
  does not maintain the trace (and is therefore compatible with the form at the Externalization interface where
  the trace is not externalized). As shown in various cases in \cite{MCB}, the interplay between these two
  forms of the coproduct plays an important role: for example in the recursive structure of 
  semantic parsing (see Chapter~3 of \cite{MCB}) the quotient $T/^d F_{\underline{v}}$ also plays a role
  as it allows for the isolation of parts of structure that are complementary to a certain accessible term.
  So it is not only $T/^c F_{\underline{v}}$ that plays a role at the syntax-semantics interface. We will see
  the same phenomenon here, where the quotient $T/^c F_{\underline{v}}$ will play a role with performing 
  movement compatibly with the assignments of theta roles, while $T/^d F_{\underline{v}}$ will be involved
  in a recursive checking that the theta criterion holds globally throughout the entire structure. 
  
  \begin{defn}\label{LBroot}
  As in Lemma~\ref{opBOalgLB}, given a tree $(c,T,\underline{c})\in  \bB_\Theta(\cM_h)$,
  we write $\wp_{in}(c,T,\underline{c})=(T, \underline{c})$. We then denote by
  $\fF(\cL(\bB_{\Theta,b}))$ denote the set of forests with
 components $(c,T,\underline{c})$ in $\bB_\Theta(\cM_h)$ such that $(T, \underline{c})\in \wp_{in}(\cL(\bB_{\Theta,b}))$,
 namely all the colors below the root vertex are as in $\cL(\bB_{\Theta,b})$.
  \end{defn}
  
 Requiring $(T, \underline{c})\in \wp_{in}(\cL(\bB_{\Theta,b}))$ rather than $(c,T,\underline{c})\in \cL(\bB_{\Theta,b})$
 allows for the possibility of a color $c$ at the root that is not possible in $\cL(\bB_{\Theta,b})$ (namely the non-theta
 color $\theta_0$), while the rest of the structure below the root is as prescribed by the building rules of $\cL(\bB_{\Theta,b})$.
 
 \begin{thm}\label{ThetaC0}
 Given $(c,T,\underline{c})$ in $\cL(\bB_\Theta)$, 
 there is an admissible cut $C_0$ of $T$ such that all the cut legs in $C_0$ have color $\theta_0$ (see Remark~\ref{colorVE}) and such that the remainder part $T/^d \pi_{C_0}(T)$ with
 the deletion quotient contains no non-theta positions. We write $\Pi_0: \bB_\Theta(\cM_h) \to \bB_{\Theta,b}(\cM_h)$ for
 the projection map $\Pi_0: T \mapsto T/^d \pi_{C_0}(T)$ that cuts the forest $\pi_{C_0}(T)$ and contracts 
 edges in the remaining non-full tree, to obtain the unique maximal full binary rooted tree 
 $T/^d \pi_{C_0}(T)$  determined by it. Inductively, perform the same kind of 
 admissible cut $C_0$ on each component of the previous cut $\pi_{C_0}(T)$, until one obtains a forest
 consisting of a collection of trees with no non-theta positions and a separate collection of disjoint leaves
 marked with either a non-terminal $\theta_0$ or a terminal $(\alpha, \theta_0)$. Extend the projection
 $\Pi_0$ to a map $\tilde\Pi_0: T \mapsto F$, where $F$ is the forest $\sqcup_i T_i /^d \pi_{C_0,i}(T_i)$
 where the $T_i /^d \pi_{C_0,i}(T_i)$ are the quotient trees obtained 
 at each step of this repeated decomposition. 
 Let $\fF(\cL(\bB_{\Theta,b}))$  be as in Definition~\ref{LBroot}.  
 Then the set of balanced theta structures (or balanced complete theta 
 structures) is the intersection $\cL(\bB_\Theta)\cap \tilde\Pi_0^{-1}(\fF(\cL(\bB_{\Theta,b})))$. The elements in
 this set are those elements of $\cL(\bB_\Theta)$ that satisfy the theta criterion.
 \end{thm}
 
 \proof 
  The admissible cut $C_0$ is obtained in the following way: starting at the root of $T$, for each leaf $\ell$,
 consider the oriented path from the root to $\ell$, and cut the first edge with target vertex marked by
 $\theta_0$, if there is one (perform no cut along that path otherwise). 
 The result is an admissible cut, as there is not more than one cut along each path from root to leaves.
 Any admissible cut $C$ gives a component of the coproduct of the form $\pi_C(T)\otimes \tilde\rho_C(T)$,
 where $\pi_C(T)$ is a forest (what has been cut off from $T$) and $\tilde\rho_C(T)$ is a non-full
 binary rooted tree (with non-branching vertices) consisting of what remains attached to the root of $T$
 after the cut $C$ is performed. The non-full tree $\tilde\rho_C(T)$ contains no non-theta vertices (marked by
 $\theta_0$) since $C_0$ cuts above any first occurrence of $\theta_0$ on paths from the root.  
The  non-full tree $\tilde\rho_C(T)$
 determines a unique maximal full binary rooted tree, the deletion quotient $\rho_C(T)=T/^d \pi_C(T)$, 
 obtained from $\tilde\rho_C(T)$ by some edge contractions. The map 
 $\Pi_0: T \mapsto T/^d \pi_{C_0}(T)$ is a map of sets $\Pi_0: \bB_\Theta(\cM_h) \to \bB_{\Theta,b}(\cM_h)$.
 It is not a morphism of operads, because it does not preserve degrees, since $T/^d \pi_{C_0}(T)$ has
 a fewer leaves than $T$, but it does satisfy a weaker compatibility with the operad compositions,
 namely if $\circ_i$ is an operad insertion $T\circ_i T'$ at one of the leaves of $T$ that are also leaves
 of $T/^d \pi_{C_0}(T)$, then $\Pi_0(T\circ_i T')=\Pi_0(T)\circ_i \Pi_0(T')$. Any tree $T$ of an element
 of $\cL(\bB_\Theta)$ is obtained from the elements in $\cR_\Theta$ 
 by repeated operad insertions. In particular, this means that all the structures generated
 using the rules \eqref{Rule1}, \eqref{Rule2}, \eqref{Rule3}, \eqref{Rule3b}, \eqref{Rule4}, with $c\neq \theta_0$, will
 result in trees with non-theta marker $\theta_0$ occurring as 
 illustrated in example \eqref{thetatheta0tree}, 
 with possibly multiple parallel $\theta_0$-marked leaves at each location where one occurs in \eqref{thetatheta0tree}.
 When performing the cut $C_0$ on a tree in $\cL(\bB_\Theta)$, one is left with a tree $T/^d \pi_{C_0}(T)$ of the form 
 $$
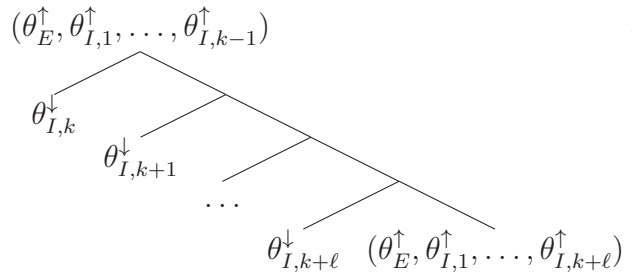
\Tree[ .{$(\theta_E^\uparrow, \theta_{I,1}^\uparrow, \ldots, \theta_{I,k-1}^\uparrow)$}  $\theta_{I,k}^\downarrow$ [  $\theta_{I,k+1}^\downarrow$ [ $\ldots$ [  $\theta_{I, k+\ell}^\downarrow$ {$(\theta_E^\uparrow, \theta_{I,1}^\uparrow, \ldots, \theta_{I,k+\ell}^\uparrow)$} ] ] ] ]  \, , $$
which is a comb tree with only theta markers, but which in general does not necessarily satisfy the theta criterion. Indeed, the
 sequence of $\theta_{I,k}^\downarrow$ in general need not completely balance the theta roles present at the insertion of 
 $\underline{\theta}^\uparrow$.  The theta criterion is satisfies in the quotient if the resulting $T/^d \pi_{C_0}(T)$ has a form
 $$ \Tree[ .$c$ $\theta_E^\downarrow$ [ $\theta_{I,1}^\downarrow$ [ $\ldots$ [ $\theta_{I, n}^\downarrow$ $\underline{\theta}^\uparrow$ ] ] ] ]  $$ 
 as in \eqref{barethetatree}.
  The $\theta_0$ markers can also occur at root vertices, which are then inserted at $\theta_0$-marked leaves
 via an operad insertion at a generator of type \eqref{Rule1}, or \eqref{Rule4} with $c=\theta_0$, or \eqref{Rule5}. 
 The resulting subtrees with $\theta_0$-marked root vertex
 will be cut off by one of the $C_0$ cuts in the recursive procedure that defines $\tilde\Pi_0$. 
 This is the reason why 
 we use the condition $(T, \underline{c})\in \wp_{in}(\cL(\bB_{\Theta,b}))$ in Definition~\ref{LBroot}.  
  Thus, in $\tilde\Pi_0(T)$ each resulting component
 $T_i /^d \pi_{C_0,i}(T_i)$ is a comb tree with only theta-markers and no $\theta_0$ marker, except possibly at the root.
 These are not necessarily generators in $\cR_b$ because they do not in general satisfy the
 theta criterion. However, those that do satisfy the theta criterion are exactly the ones that map under $\wp_{in}$
 to elements $\wp_{in}(\cR_b)$ (that is, are in $\cR_b$ except for possibly a $\theta_0$ marker at the root vertex). 
 Thus $T$ satisfies the theta criterion if all the quotients $T_i /^d \pi_{C_0,i}(T_i)$
 of the repeated application of the admissible cut at the first occurrences of $\theta_0$ on paths
 from the root will be all of the form \eqref{barethetatree} satisfying the theta criterion (including
 the possibility $c=\theta_0$). 
 \endproof

\begin{rem}\label{roles}{\rm In the description $\cL(\bB_\Theta)\cap \tilde\Pi_0^{-1}(\fF(\cL(\bB_{\Theta,b})))$ of the 
set of complete theta structures, the set $\cL(\bB_\Theta)$ describes the relation of theta roles to argument structure,
the non-theta positions (including possible arbitrarily long chains of adjunctions), the external and internal arguments, 
and the valence of predicates. The set $\tilde\Pi_0^{-1}(\fF(\cL(\bB_{\Theta,b})))$ describes the matching of theta
roles according to the theta criterion. The rule \eqref{Rule5} is the one that accounts for the dichotomy between 
External and Internal Merge in semantics and the fact that movement by Internal Merge is to a non-theta
position, as we discuss in \S \ref{EMIMsemSec} below, see also Remark~\ref{M1}. 
}\end{rem}

We discuss in the next section how the special form of the theta-structures \eqref{barethetatree} and
\eqref{thetatheta0tree} are compatible with movement and with other resulting structures.

\section{The Internal and External Merge dichotomy in Semantics} \label{EMIMsemSec}

In this section we discuss movement (Internal Merge) and the effect it has on the coloring
algorithms discussed in the previous section. As we have seen, the coloring that assigns theta 
roles and checks for the theta criterion is based on the algebraic structure of colored operads
and their generating systems, and the notion of algebra over an operad. These are 
algebraic structures on binary rooted trees different from the structure of Hopf algebra
that accounts for structure formation via Merge (though we did use the coproduct to 
implement recursively the checking for the theta criterion). Thus, we seem to have two different
viewpoints here:
\begin{enumerate}
\item Free structure formation via Merge using the Hopf algebra structure on workspaces, and then a separate mechanism to 
implement coloring (assignment of theta roles) and filter out structures without valid coloring and/or not
satisfying the theta criterion.
\item Structure formation via Merge that already accounts for coloring, while structure is formed, and directly 
builds only structures with valid coloring. The theta criterion still has to be implemented as a filter on
fully formed structures (at least at the level of phases), as otherwise it may be satisfied locally but violated globally.
\end{enumerate}
The recent literature on free symmetric Merge sometimes emphasizes one of these
viewpoints (as in \cite{MCB}) or the other (as in \cite{ChomskyElements}). In fact, as
we are going to show here, these two formulations are entirely equivalent, so it is only a matter of convenience to
phrase theta theory in one or the other framework. In particular, while in the previous
sections we took the first viewpoint, in this section, in order to discuss movement and
the dichotomy in semantics, it will be more convenient to adopt the second.

\subsection{The dichotomy in semantics} \label{dichoSec}

It was briefly discussed in Section~3.8.1 of \cite{MCB} how one sees in the mathematical setting
the dichotomy (often referred to as ``duality")
in Semantics, between  External  and Internal Merge, pertaining to propositional domain (theta-structure) 
and clausal domain (information-related), respectively. As described, for instance, in \S 5.2 of \cite{ChomskyElements},
this dichotomy postulates a segregation of EM and IM, where only 
External Merge accounts for argument structure and $\theta$-role assignments, while
Internal Merge maintains a separate role, related to non-argument structure and 
information-related properties. In particular movement produced by IM cannot be to
a theta-position. 

In the setting of SMT, where the core computational model of syntax consists of structures
freely formed by a symmetric Merge (arbitrary non-planar binary rooted trees with leaves 
labelled by lexical items and syntactic features), constraints barring movement by IM
to theta-positions are imposed, like other constraints (head and complement structure,
labeling algorithms, phase theory, matching of theta role assignments) in the forms of
filters on the freely formed structures that eliminate those not satisfying the constraints.

In the case of the EM/IM dichotomy, it is observed in Section~3.8.1 of \cite{MCB} that this
is visible directly in the model, from the fact that the structure of syntactic objects
as algebra over an operad, which is viewed as the key underlying structure to theta theory,
is based on EM (or more precisely on the magma operation on syntactic objects), hence
IM is absent from the building of theta relations.

This captures one aspect of the dichotomy, but it does not account directly for another
aspect, which is the barring of IM movement to theta positions. We discuss it here. 

\subsubsection{The Merge action} \label{MergeActSec}
As we recalled in \S \ref{coprodSec} above, we can formulate the action of Merge on
workspaces in the form of a linear operator on the vector space $\cV(\fF_{\cS\cO_0})$
spanned by the workspaces, namely the forests $F=\sqcup_a T_a$ with components
that are syntactic objects $T_a\in \fT_{\cS\cO_0}$, with
\begin{equation}\label{Merge}
 \fM_{S,S'}(F) =\sqcup \circ (\cB \otimes {\rm id}) \circ \delta_{S,S'} \circ \Delta\, , 
\end{equation}  
where $\Delta$ is the coproduct on $\cV(\fF_{\cS\cO_0})$ (defined in terms of
admissible cuts), $\delta_{S,S'}$ is the projection on the components of the
coproduct that have $S\sqcup S'$ in the left channel, $\cB$ is the grafting operator
that grafts the root vertices of a forest at a common root, and $\sqcup$ is the product
on workspaces given by disjoint union. As discussed in \cite{MCB}, Internal Merge
can be represented in the form $\fM_{T_v,T/T_v} \circ \fM_{T_v,1}$, where $T$ is
a component of the workspace $F$ and $T_v$ is an accessible term of $T$ (see
\S 1.4.3 of \cite{MCB}). On the other hand, External Merge takes the form
$\fM_{T_a,T_b}$ where $T_a,T_b$ are two different components of the workspace $F$.
The formulation above also accounts for three forms of Sideward Merge, respectively of the form 
$\fM_{T'_w,T}$ with $T$ a component of $F$ and $T'_w\subset T'$ an
accessible term of a different component, $\fM_{T_v,T'_w}$ with $T_v\subset T$
and $T'_w\subset T'$ accessible terms of two different components, and
$\fM_{T_v,T_w}$ with $T_v\sqcup T_w \subset T$, disjoint accessible terms 
of the same component $T$ of the forest $F$. 

When one considers all the possible Merge operations that can be performed on
a given workspace, one obtains an operator of the form 
$$ \cK = \sum_{S,S'} \fM_{S,S'} = \sum_{S,S'} \sqcup \circ (\cB\otimes {\rm id}) \circ \delta_{S,S'}\circ  \Delta $$
which is a Hopf algebra Markov chain in the sense of \cite{Diac}.

\subsubsection{Internal Merge and non-theta positions}\label{secIMnontheta}

The formulation of Merge as a Hopf algebra Markov chain, as recalled above, 
makes the property that IM does not move to a theta position a subtle point, 
which, at first, may appear difficult to account for in the model. The difficulty comes from the fact that,
stated in these terms, this appears to be a constraint on the action of the free 
symmetric Merge. More precisely, one can view it in this way: the property that
IM does not move to a theta position appears to delete some of the components
$K_{F,F'}$ of the Hopf algebra Markov chain given by the action of free
symmetric Merge on workspaces $F,F'$ (see Section~1.9 of \cite{MCB}), in a way
that is not the result of a global filter (projection $\Pi$ on the vector space
generated by workspaces), namely it's not obtainable as 
$\cK=(K_{F,F'})_{F,F'} \mapsto \Pi \cK \Pi$. This kind of global projection $\Pi$
can be seen as a way of filtering out some of the structures freely formed by Merge,
after structure formation is achieved (for example, implementation of syntactic
parameters in Externalization). On the other hand, a direct modification of the
components $K_{F,F'}$ of $\cK$ would be violating the principle of SMT, that 
free symmetric Merge is the only structure formation process of syntax.

This apparent problem can in fact be resolved within the mathematical model of
Merge and SMT of \cite{MCB},
compatibly with the SMT assumption that free structure formation is not
constrained. Indeed, the constraints that implement the EM/IM dichotomy
can be expressed entirely in terms of {\em constraints on coloring rules} without
imposing any constraints on structure formation, as we show in the rest of this section. 

\subsection{Filter by coloring}\label{FilterSec}

The implementation of the assignment and matching of theta roles can
be seen as a filtering of the structures freely generated by the free
symmetric Merge acting on workspaces (modeled as in \cite{MCB}, see \S \ref{MergeActSec} above).
Given a syntactic object $T\in \fT_{\cS\cO_0}$, a first filtering checks
for a well defined syntactic head $h_T$. If $T\in {\rm Dom}(h)$, then
the filtering by theta roles starts by examining the lexical items and
syntactic features at the leaves $\ell$ that are heads that project according
to the head function $h_T$, and the associated theta-grid data $\underline{\theta}_\ell^\uparrow$.
These contain all the theta roles that need to be assigned matching the
argument structure of elements in $\cL(\bB_\Theta)$ and the
theta criterion of elements in $\tilde\Pi_0^{-1}(\fF(\cL(\bB_{\Theta,b})))$
(see Theorem~\ref{ThetaThetab} and Remark~\ref{roles}). If the colors
$\underline{\theta}_\ell$ can propagate to a coloring of all vertices of $T$
that is compatible with the coloring rules resulting in an element
of $\cL(\bB_\Theta) \cap \tilde\Pi_0^{-1}(\fF(\cL(\bB_{\Theta,b})))$
(a balanced complete theta structure) then the syntactic object is kept and
available for mapping to the syntax-semantics interface and to
Externalization, or else it is filtered out. With this description of
the checking of theta roles, no constraint is imposed on the
action of the free symmetric Merge.

\smallskip

However, the result of this filtering is equivalent to a reformulation
that can be thought of as a constraint on the action of Merge, as we now
illustrate. Since this is only an equivalent rephrasing, the main point
of these two different interpretations is that, by viewing the theta theory
as filters on fully formed structures, one can completely decouple
structure formation (free symmetric Merge acting on workspaces)
from filters (head and complement structure, theta theory, parameters).
This decoupling is the main theoretical and conceptual advantage of the
SMT model. It allows for simpler arguments dealing separately with
each modules of the faculty of language. On the other hand, the other
equivalent formulation that we now discuss, that views this filtering
as a constraint on Merge, has the advantage of cutting down the
combinatorial explosion of the Merge action, in the same way in
which phases (which a priori are also a later filter coming from 
the head and complement structure and the labeling algorithm)
can be also thought of equivalently as a constraint on the Merge
action, which adapts the coproduct to phases and reducing the
combinatorial explosion. (This case of phases is discussed in Sections~1.13, 1.14,
and 1.15 of \cite{MCB}.) The main point in this reinterpretations of
filters as Merge constraints is just the observation that the decoupling
of structure formation from filtering does not require that they happen in
a strict chronological order in the process of language generation. Rather,
given that structure formation is a bottom-up process that builds smaller
structures first in intermediate workspaces and then gradually combines
them into larger structures, the filtering can happen immediately after
any structure is formed, thus eliminating any that does not pass the
filter conditions before it is combined into larger structures.
This intertwining of structure formation and filters is equivalent
to the decoupling that separately analyzes structure formation
and filters as separate parts of the model, but at the same time
it cuts down the combinatorial explosion by reducing the number of
structures as they are being built. We can describe the filter of
theta theory equivalently as a colored version of the structure
formation process, which is a way of viewing the filtering as
happening at each stage of the structure formation. Since these
two descriptions are equivalent, this is only a choice of convenience:
while full decoupling is better for theoretical questions (proving results
about various parts of the system), intertwining of filters and
structure formation is better for computational and algorithmic
aspects, and especially for computer implementation, where
containing the combinatorial explosion becomes more important. 

\subsection{Build by coloring} \label{BuildColorSec}

We show here that the dichotomy in semantics between External and Internal Merge
is a consequence of the coloring description of theta roles, when the latter is reinterpreted
equivalently as a colored version of the structure formation.

Usually, this dichotomy is described purely in terms of External and Internal Merge.
The Sideward Merge components of the Merge
action on workspaces are subdominant with respect to External and Internal
Merge, in terms of cost optimality with respect to Minimal Search and
Minimal Yield (Resource Restriction). However, there are reasons to expect a limited
presence of SM operations, both from the theoretical perspective (based on the
analysis of the properties of Merge as a Hopf algebra Markov chain) and
from the empirical linguistics perspective where certain phenomena
seem best explained by a limited form of Sideward Merge (see the
discussion in \cite{MarLarHuij}). Thus, it is a relevant question what behavior
these should have with respect to the EM/IM dichotomy. We will answer this
question in Corollary~\ref{SMtheta} below.

\begin{defn}\label{setXi}
Let $\Xi_{\bB_\Theta}$ be the set of pairs $(c, \{ c', c'' \})$ of $c\in \cI_\Theta$ and $c',c''\in \Theta$ with
the property that, for $T$ the cherry tree with three vertices (a root and two leaves) and two edges,
the element $(c, T, (c',c''))$ in $\bB_\Theta(\fM_h)(2)$ is a generator in $\cR_\Theta$.
\end{defn}

\begin{defn}\label{ColorMerge}
For $c\in \cI_\Theta$, the grafting operator $\cB^c$ is the operator that
takes a forest $F=T_1\sqcup \cdots \sqcup T_N$ with colored vertices and maps
it to the rooted tree
$$ \cB^c(F)=\Tree[.$c$ $T_1$ $T_2$ $\cdots$ $T_N$ ] $$
When $F=T \sqcup T'$ consists of two components then 
$$ \cB^c(F)=\Tree[.$c$ $T$ $T'$ ] =: \fM^c(T,T') \, . $$
\end{defn}

\begin{lem}\label{McolorLem}
The {\em colored} Merge operators $\fM_{S,S'}^c$ is given by
\begin{equation}\label{Mcolored}
\fM^c_{S,S'}=\sqcup \circ (\cB^c \otimes {\rm id}) \circ \delta^c_{\{ c_S, c_{S'} \}} \delta_{S,S'} \circ \Delta \, ,
\end{equation}
where $\delta^c_{\{ c_S, c_{S'} \}}=1$ for $(c, \{ c_S, c_{S'} \}) \in \Xi_{\bB_\Theta}$ and
$\delta^c_{\{ c_S, c_{S'} \}}=0$ for $(c, \{ c_S, c_{S'} \}) \notin \Xi_{\bB_\Theta}$. The $\fM_{S,S'}^c$ 
define linear operators on the vector space $\cV(\fF(\cL(\bB_\Theta)))$ spanned by
workspaces where all the components are in $\cL(\bB_\Theta)$.
\end{lem}

\proof The $\delta^c_{\{ c_S, c_{S'} \}}$ selects only the components where the colors of
the root vertices of $S$ and $S'$ and the color $c$ of the new root vertex are compatible with
one of the rules \eqref{Rule1}--\eqref{Rule5} in $\cR_\Theta$ as in Definition~\ref{ThetaBudSys}.
Note that all the rules in $\cR_\Theta$ for the theta bud system $\bB_\Theta$ are coloring
rules on a single binary Merge operations, so the $\delta^c_{\{ c_S, c_{S'} \}}$ capture all the
generating system for the language $\cL(\bB_\Theta)$. This implies that, if $F\in \fF(\cL(\bB_\Theta))$,
hence $S,S' \in \cL(\bB_\Theta)$, then $\fM^c_{S,S'}(F)\in \cV(\fF(\cL(\bB_\Theta))$, since
$\cB^c(S\sqcup S')$ is an operad composition of elements in $\cR_\Theta$, it is again in $\cL(\bB_\Theta)$.
\endproof

Note that in the physics literature the grafting operator $\cB$ is also always considered
in a form $\cB^c$ with a choice of coloring of the root vertex, and the universal property
that the Hochschild cocycle $\cB^c$ satisfies also depends on having a collection of
$\cB^c$ operators with different coloring, see for instance the discussion in \cite{Panzer}.

The fact that structure building through the action of the colored Merge operators $\fM^c_{S,S'}$ on
the space $\cV(\fF(\cL(\bB_\Theta)))$ of workspaces, followed by checking theta criterion by
projecting onto $\tilde\Pi_0^{-1}(\fF(\cL(\bB_{\Theta,b})))$,
is equivalent to considering all the structures formed by the ordinary Merge of \eqref{Merge} on the
space of all workspaces, followed by filtering by projection onto 
$\cV(\fF(\cL(\bB_\Theta)))\cap \tilde\Pi_0^{-1}(\fF(\cL(\bB_{\Theta,b})))$, can be seen
as the combination of two facts:
\begin{itemize}
\item By Lemma~\ref{McolorLem}, the coloring of the Merge operation is equivalent to filtering 
according to unbalanced theta structures (belonging to the set $\cL(\bB_\Theta)$) 
after each step of structure formation by action of Merge on workspaces, followed by filtering by theta criterion 
(beloning to the set $\tilde\Pi_0^{-1}(\fF(\cL(\bB_{\Theta,b})))$ on the final structure, or on each intermediate 
completed phase.
\item The checking for $\cL(\bB_\Theta)$ after each step of structure formation and for 
$\tilde\Pi_0^{-1}(\fF(\cL(\bB_{\Theta,b})))$ after each completed phase is equivalent
to checking for $\cL(\bB_\Theta) \cap \tilde\Pi_0^{-1}(\fF(\cL(\bB_{\Theta,b})))$ on the
final resulting structure. 
\end{itemize}

\begin{prop}\label{IMnotheta}
It is a consequence of the coloring algorithm of the theta bud system that
Internal Merge can only move to non-theta positions.
\end{prop}

\proof Internal Merge is of the form $\fM_{T_v,T/T_v} \circ \fM_{T_v,1}$, where $T$ is
a component of the workspace $F$ and $T_v$ is an accessible term of $T$, see
\S 1.4.3 of \cite{MCB}. Thus, in the colored form, it is of the form
$\fM^{c'}_{T_v,T/T_v} \circ \fM^{\theta_0}_{T_v, (1,\theta_0)}$. This means that 
$\fM^{\theta_0}_{T_v, (1,\theta_0)}(T)$ is the tree $T_v$ with the root colored by
$\theta_0$. The application of $\fM^{c'}_{T_v,T/T_v}$ produces a term $\fM^{c'}(T_v, T/T_v)$,
where the root of the entire structure is labelled $c'$ and the root of the
substructure $T_v$ is labelled $\theta_0$, which means that the accessible term $T_v$
that is moved by IM moves to a non-theta position, while the whole structure 
$\fM^{c'}(T_v, T/T_v)$ can attach to a $c'$ position of a larger structure.
\endproof

\begin{prop}\label{EMtheta}
The fact that all the assignments of theta roles are performed by External Merge
is a consequence of the coloring algorithm of the theta bud system and of Lemma~\ref{McolorLem}.
\end{prop}

\proof The statement is equivalent to showing that the language $\cL(\bB_\Theta)$ is obtained
by repeated application of colored External Merge, starting from elements of $\cR_\Theta$
with terminal colors at the leaves. This is clear by Lemma~\ref{McolorLem}, since repeated
application of colored External Merge inductively consists of colored operad compositions 
of two previously constructed elements (with terminal leaves) inserted in the two non-terminal
leaves of one of the elements of $\cR_\Theta$ with non-terminal colors. 
Note that $\cL(\bB_\Theta)$ includes assignments of theta roles that are not
balanced by the theta criterion, while the latter corresponds to intersecting with 
$\tilde\Pi_0^{-1}(\fF(\cL(\bB_{\Theta,b})))$. However, since all of 
$\cL(\bB_\Theta)$ is obtainable by colored External Merge, so is the
subset where the theta criterion holds.
\endproof

Finally, we can also address the question of how the Sideward Merge components
of \eqref{Merge} behave with respect to the EM/IM dichotomy in semantics. 

\begin{cor}\label{SMtheta}
Sideward Merge behaves like External Merge and not like Internal Merge
in the dichotomy, namely it can move extracted accessible terms to theta positions, 
unlike Internal Merge.
\end{cor}

\proof This is a direct consequence of the colored Merge operation \eqref{Mcolored}
applied to the three cases of Sideward Merge: $\fM_{T'_w,T}$, and $\fM_{T_v,T'_w}$,
and $\fM_{T_v,T_w}$. In all of these cases, when applying the colored merge, the
root vertices of the extracted accessible terms maintain the color they had in the
original component, which can be either a theta or a non-theta position, unlike the
case of Internal Merge where the color of the root of the extracted accessible term
is always changed to $\theta_0$ because the extraction is performed through 
an operator $\fM_{S,1}$.
\endproof

\subsection{Effects of movement} \label{MoveTheta} 

As a consequence of the formulation of colored Merge described in this section,
and in particular the behavior of Internal Merge moving to non-theta positions,
we can see that the structure of theta roles in specific examples. 

\begin{ex}\label{Ex1}{\rm
The theta roles structure of a sentence like
{\em Mary gave a book to John} 
is not just given by a comb tree of the form
\begin{equation}\label{nocomb}
 \Tree[ $\theta_E^\downarrow$ [ $\underline{\theta}^\uparrow$ [ $\theta_{I,1}^\downarrow$ $\theta_{I,2}^\downarrow$ ] ] ]  
\end{equation} 
with $$\underline{\theta}^\uparrow=(\theta_E^\uparrow, \theta_{I,1}^\uparrow, \theta_{I,2}^\uparrow)=((\text{agent}_E^\uparrow, \text{theme}_{I,1}^\uparrow, \text{goal}_{I,2}^\uparrow)\, , $$
but it is obtained via IM movement from a comb tree of the form
$$ \Tree[ $\theta_E^\downarrow$ [ $\theta_{I,1}^\downarrow$ [ $\theta_{I,2}^\downarrow$ $\underline{\theta}^\uparrow$ ] ] ]  $$
which gives the tree structure 
$$ \Tree[ .$c$  {(Mary, $\theta_E^\downarrow$)}  [ .$\theta_E^\uparrow$  {(gave, $\theta_0$)} [  .$\theta_E^\uparrow$  {(a book, $\theta_{I,1}^\downarrow$)}  [ .$\underline{\theta}_1^\uparrow$  {(to John, $\theta_{I,2}^\downarrow$)}    {$\underline{\theta}^\uparrow$}   ] ] ] ] $$ 
where IM has moved a leaf marked by (gave, $\underline{\theta}^\uparrow$) to the non-theta position 
(gave, $\theta_0$), while $\underline{\theta}^\uparrow$ remains as coloring of the trace of the movement. 
This description in terms of movement is compatible with what described in \cite{Larson}. }
\end{ex} 

Here it is important that we use the contraction form of the coproduct $T/^c T_v$, since we need to
have the trace of the movement so that the original coloring at the roots of the extracted substructure $T_v$
is still present at the trace leaf, while the new position where IM moves the accessible term $T_v$ to is marked by
a non-theta $\theta_0$. 

\begin{ex}\label{Ex2} {\rm 
A similar example occurs, for instance, with the oblique object in Mandarin. Using an example
discussed in \cite{LarZha}, the structure that is obtainable in our model of theta roles is given by
$$ \Tree[ .$c$  {(Zhangsan, $\theta_E^\downarrow$)}  [  .$\theta_E^\uparrow$ {(song (gei), $\theta_0$)} [ .$\theta_E^\uparrow$ {(Lisi, $\theta_0$)}  [ .$\theta_E^\uparrow$  {(kuai qian, $\underline{\theta}_1^\downarrow$)} [  .$\underline{\theta}_1^\uparrow$  $\underline{\theta}_2^\downarrow$  $\underline{\theta}^\uparrow$ ] ] ] ] ] $$   
where in this case two IM operations move terms to non-theta positions. This is compatible with
the movement description of the derivation of this sentence given in \cite{LarZha}. }
\end{ex}

In these examples
the trace remains the position where the theta roles are injected into the structure (the theta giver position) even if
the actual predicate that assigns the theta roles has moved under the effect of IM, which in our formulation can
only move to a $\theta_0$ location because of the Merge with $(1,\theta_0)$ involved in the IM operation. What makes IM special with respect to its behavior in theta theory (and different from both EM and SM, as we have seen) is exactly the fact
that IM is a composition with a $\fM_{S,1}$ operation involving the unit of the magma, where $1$ can only
be colored by $\theta_0$. If IM were to move to a theta position, then the only choice would be that
it moves with a theta label equal to the one in its original location.That would then cause the failure of the theta
criterion, even if the structure one starts with satisfies it. Indeed, since the trace of the movement exists at the
CI interface, the trace would also have to maintain the coloring it already has at the root of the extracted term $T_v$
(there is no role for uncoloring or for changing color to an already colored vertex that is not moved). but them
the same theta role would be counted twice and no longer matched in the theta criterion. Thus, indeed this
agrees with the usual formulation of the dichotomy in semantics. 

Note that, in principle, it would be possible to account for the theta roles assignments in cases like
Example~\ref{Ex1} and \ref{Ex2}, by introducing more generators and use a larger colored operad, for
example by having additional generators of the form \eqref{nocomb} and other similar structures with
the insertion of the theta roles $\underline{\theta}^\uparrow$ happening at different positions in the tree.
This would avoid having to use movement to non-theta positions for the giver of the theta roles, with
the insertion of the theta roles remaining at the trace position. 
However, the formulation given here is the minimal one, in terms of generators used, and increasing
the number of generators leads to more structures, several of which then would
have to be filtered out as non-viable.  The more restrictive formulation introduced here seems
to account for the viable structures with a minimal generating set. 
  
\section{Refinements of the model}

The choice presented here of generating system for the colored operad of theta relations
is designed to minimize the number of generators needed, and to account for other
configurations of theta positions via movement, namely moving via IM to non-theta positions
and maintaining the theta role insertion at the leaf with the trace of the movement, as
discussed in \S \ref{MoveTheta}. This choice has some possible drawbacks, which have
to do with constraints of compatibility between the structure of theta relations and the
structure of head and complement and phases. Although we will not discuss the relations
between these two structures explicitly in this paper, we can illustrate here briefly some of the
issues involved, to which we will return in future work. 

\subsection{The external argument}\label{extargprobSec}

As observed in Remark~\ref{gridrem}, our theta grids $\underline{\theta}=(\theta_E, \underline{\theta}_I)$ include
the external argument, hence the insertion point that has the theta-color $\underline{\theta}^\uparrow$ is also assigning
the $\theta_E$ external argument, not only the $\theta_{I,k}$ internal arguments. This choice has the advantage
that it provides us with colored operad generators that satisfy a ``local conservation" of theta role at each node,
namely all the generators \eqref{Rule1}, \eqref{Rule2}, \eqref{Rule3}, \eqref{Rule3b}, and \eqref{Rule4} satisfy a
local form of the theta criterion, which can be phrased as the property that
\begin{itemize}
\item All colored operad generators not involving the $(1,\theta_0)$ color satisfies a conservation of theta roles,
namely any theta role $\theta^\uparrow$ that is flowing into the vertex from one of the leaves
is either flowing out at the root in the form of the same $\theta^\uparrow$ or is being discharged at the other
leaf as a $\theta^\downarrow$.
\end{itemize}
The case of the $(1,\theta_0)$ color is the only exception where the $(1,\theta_0)$ can absorb a theta color
and deliver $\theta_0$ at the root.

It is a natural question, then, how one can reconcile this formulation, where the $\theta_E^\uparrow$ is
inserted at the same head where all the other $\theta_{I,k}^\uparrow$ are inserted, with the usual formulation
in Minimalism that separates the assignment of the external theta role from the internal ones, with the
$\theta_E$ assigned by $v^*$. Namely, the external argument of a phrase is one that receives 
a theta-role independently of the phrasal head. For a discussion of this point within the most recent
formulation of Minimalism, see \cite{ChomskyGK} and \cite{ChomskyElements}, where the $\theta_E$
role is Merged with External Merge outside of the $v^*P$ phase, so that the EA receiver of $\theta_E^\downarrow$
is in the CP phase but not in the $v^*P$ phase, and receives its $\theta_E^\downarrow$ theta role from 
$v^*P$, the projection of the $v^*$ head, so that $v^*$ plays the role of the $\theta_E^\uparrow$ giver. 

For example, consider the case of the (non-planar) tree
$$ T= \Tree[ .$v^*P$  Alice [ .$v^*P$  $v^*$ [ .$VP$  Fido fed  ] ] ] $$
With the formalism we described in the previous sections, we would regard its
theta structure as obtained by the coloring $(c,T,\underline{c})$ arrived at by 
decomposing $T$ into three operad insertions of generators in $\cR_\Theta$.
Given the generating system $\cR_\Theta$, the bottom generator in this case has to be of the form
$$ 
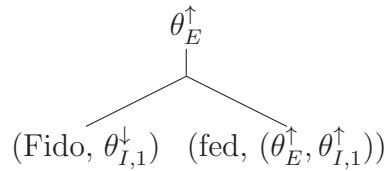
\Tree[ .$\theta_E^\uparrow$ [ {(Fido, $\theta_{I,1}^\downarrow$)} {(fed, $(\theta_E^\uparrow,  \theta_{I,1}^\uparrow)$)} ] ] 
$$
and the top one has to be of the form
$$ 
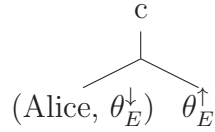
\Tree[ .c [ {(Alice, $\theta_E^\downarrow$)} $\theta_E^\uparrow$  ] ] 
$$
where the color $c$ at the root would be $\theta_0$ unless the whole structure participates
into some higher structure where it either assigns or receives a theta role.
This in turn uniquely determine which generator can fit in the intermediate position: it has to be of the form
$$ 
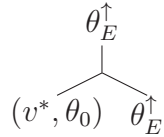
\Tree[ .$\theta_E^\uparrow$ [ {$(v^*, \theta_0)$} $\theta_E^\uparrow$ ] ] 
$$
since this is the only generator that can have the same theta giver on one of the leaves and at the root.
Thus, with this form of theta-coloring, the leaf $v^*$ is colored by a non-theta relation $\theta_0$, which
contrasts the idea that this $v^*$ should be associated to a $\theta_E^\uparrow$.  

However, there is a sense in which, also in this setting, we can say that the $v^*$ is
associated to a $\theta_E^\uparrow$ role. We can view the tree above in the form
$$ \Tree[ {(Alice, $\theta_E^\downarrow$)} [ .$\theta_E^\uparrow$ $v^*$ [ .$\theta_E^\uparrow$ {(Fido, $\theta_{I,1}^\downarrow$)} {(fed, $(\theta_E^\uparrow,  \theta_{I,1}^\uparrow)$)} ] ] ] \, . $$
If we interpret the color label at a vertex as coloring the edge immediately above that vertex, then we can
view $v^*$ as branching from an $\theta_E^\uparrow$-labelled edge, or yet equivalently as an edge
labelled by a pair $(v^*, \theta_E^\uparrow)$. Thus, one can reinterpret this coloring as involving
a $v^*$ in the role of a giver of the external theta role. 

Note that this interpretation only applies to the external theta role. For example, if one tries to follow
a similar reasoning with a tree drawn in the form 
$$ \Tree[ {(Alice, $\theta_E^\downarrow$)} [ .$\theta_E^\uparrow$ $v^*$ [ .$\theta_E^\uparrow$ {(Fido, $\theta_{I,1}^\downarrow$)}  [ .{$(\theta_E^\uparrow,\theta_{I,1}^\uparrow)$} $v^+$ [ .{$(\theta_E^\uparrow,\theta_{I,1}^\uparrow)$} {(to John, $\theta_{I,2}^\downarrow$)}  {(gave, $(\theta_E^\uparrow,\theta_{I,1}^\uparrow, \theta_{I,2}^\uparrow)$)} ] ] ] ] ] $$
one can no longer interpret $v^+$ as giver of a theta role, since the edge it branches from is carrying both a $\theta_E$
and a $\theta_{I,1}$. In that sense it is only the external theta role that can be also interpreted as being carried by $v^*$.
There remains a difference with respect to the usual formulation in which one thinks of the external argument as being
receiving its theta-role {\em independently} of the phrasal head, since here the $\theta_E^\uparrow$ is in any case already
injected at the phrasal head. 

A way to formulate the theta assignments that would more closely match this idea can be illustrated as follows. 
Consider again the first tree discussed
here. If we inject only the internal theta role at the phrasal head and the external one at $v^*$, resulting in 
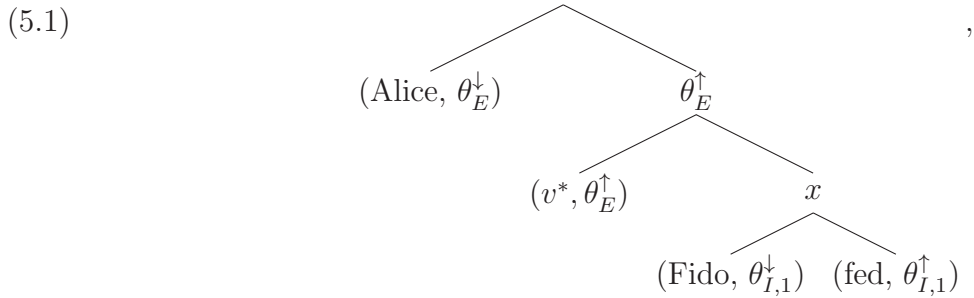
\begin{equation}\label{AliceEx2}
 \Tree[ {(Alice, $\theta_E^\downarrow$)} [ .$\theta_E^\uparrow$ {$(v^*, \theta_E^\uparrow)$} [ .$x$ {(Fido, $\theta_{I,1}^\downarrow$)} {(fed, $\theta_{I,1}^\uparrow$)} ] ] ] \, , 
\end{equation}  
then the question is how one should color the vertex $x$? Assigning to it a theta role marking would violate the
local conservation law (the local form of the theta criterion mentioned above), so the only possibility would be
to assign $x=\theta_0$. This fits the model we described, by simply dropping the requirement that if the 
head injects a single theta role, that must be the external one. This would allow for a generator like the
one at the bottom of this tree, namely a generator 
$$ \Tree[ .$\theta_0$ [ $\theta_{I,1}^\downarrow$, $\theta_{I,1}^\uparrow$ ] ] \ \ \ \text{ in addition to  } \ \ \ 
\Tree[ .$\theta_0$ [ $\theta_E^\downarrow$, $\theta_E^\uparrow$ ] ] \, . $$
This would completely disentangle the theta structure of the $VP$ phrase from the
completed theta structure of the entire phrase, with the assignment of $\theta_E$ completely decoupled from
the assignment of the $\theta_I$ roles. A drawback of making this choice would be that one can no
longer model the fact that a given verb is typically associated with a limited set of subject theta-roles: for
example, in the case above ``fed"  is associated with {\em agent} but not with {\em experiencer} as the
external theta role, so injecting $\theta_E^\uparrow =\text{agent}_E$ (rather than $\theta_E^\uparrow =\text{experiencer}_E$)
with the verb provides the correct coloring, while the complete decoupling in \eqref{AliceEx2} would incorrectly result
in accepting  both $\text{experiencer}_E$ and $\text{agent}_E$ in \eqref{AliceEx2}.

\subsection{Movement and phases}\label{movephaseSec}

As discussed in \S \ref{MoveTheta}, other sentence structures where the predicate that
assigns the theta roles is located at a different position in the tree with respect to the
case of \eqref{barethetatree} (for example cases that look like \eqref{thetaother}) 
are not directly obtainable from the colored operad generators but are explained
by movement, where the theta role insertion remains at the trace of the movement
but the lexical item associated to it is relocated by Internal Merge to a $\theta_0$-marked
position, so that the counting of theta roles matching in the theta criterion is unchanged.
This type of IM movement explanation has the advantage of accommodating cases like
certain constructions of oblique objects in Mandarin where the theta configurations are
more difficult to obtain otherwise, without having to include multiple additional generators
to the colored operad, which we try here to minimize. 

However, this formulation allows cases of movement by Internal Merge within the
interior of a phase, while in the present theory of Minimalism, 
Internal Merge only raises to the edge of the phase, in Spec-phase position, never within
the interior of a phase, and a phase impenetrability condition further prevents movement 
beyond the edge of phase.  

This means that the colored operad that accounts for the assignments of  theta roles
should interface with the structure of head and complement, and of phases, which can
itself be formulated as a coloring problem. 

Just as we achieve here the balance of theta roles under the theta criterion through
a relation between two colored operads (the bare theta structures and the extended
unbalanced ones), one can implement constraints on movement through a relation
between the colored operad of theta roles and the coloring of the phase structure. This
would force the inclusion of additional generators in the colored operad of theta
roles, to avoid IM movement that is not compatible with the phase structure, 

One can consider the overall picture in the following way: the process of free
structure formation produced by Merge is accompanied, within I-language
(prior to Externalization and further filtering by syntactic parameters) by two
fundamental filters, both implementable in the form of a coloring algorithm:
one that checks for a viable structure of head, complement, and phases, and
one that check for a good assignment of theta roles satisfying the theta condition. 
These two coloring algorithms, which in the formulation given in this paper are
seen as essentially independent of each other, and in fact much more closely
related, and the actual final form of the set of generators of the colored operad
of theta roles should be modified from the simple form described in this paper
to accommodate the more precise relation with the coloring by head, complement,
and phases. 

We will discuss the coloring formulation of head, complement, and phases elsewhere,
and we will return to discuss how this relates to the colored operad of theta roles
and extensions of its set of generators.

\bigskip
\subsection{Acknowledgment} The first author is supported by NSF grand DMS-2104330. 
We thank Riny Huijbregts for several helpful comments.

\end{document}